\documentclass{article}
\PassOptionsToPackage{inline}{enumitem}
\usepackage[arxiv1col]{paperkit}

\usepackage{amsmath,amsfonts,bm}

\def\eqref#1{equation~\ref{#1}}

\def\1{\bm{1}}

\DeclareMathAlphabet{\mathsfit}{\encodingdefault}{\sfdefault}{m}{sl}
\SetMathAlphabet{\mathsfit}{bold}{\encodingdefault}{\sfdefault}{bx}{n}

\makeatletter\renewcommand{\pk@adddot}[1]{#1}\makeatother

\usepackage{fontawesome5}
\usepackage{soul}
\usetikzlibrary{arrows.meta,positioning,fit,calc}
\usepackage{siunitx}
\newcommand{\oeb}{\textsc{OEB}}
\newcommand{\nRuns}{119}
\newcommand{\nTasks}{12}
\newcommand{\nPtbRuns}{91}

\newcommand{\nSpeedRuns}{8}
\newcommand{\nChipTasks}{5}
\newcommand{\nLabelled}{90}
\newcommand{\nCheat}{11}
\newcommand{\nCheatCaught}{10}

\newcommand{\nCaughtMedian}{3}

\newcommand{\callsMedian}{652}

\newcommand{\hmIdeas}{36}
\newcommand{\hmRateGLM}{1.6}
\newcommand{\hmRateKimi}{0.8}
\newcommand{\hmRateGPTsix}{0.3}
\newcommand{\hmRateGPTfive}{0.1}
\newcommand{\hmIdeasGLMKimi}{24}
\newcommand{\hmDeriveGLMKimi}{22}
\newcommand{\hmIdeasGPT}{4}
\newcommand{\hmLuckyGPT}{3}

\newcommand{\hmIdeasOpus}{4}
\newcommand{\hmDeriveOpus}{4}
\newcommand{\hmTasks}{5}

\newcommand{\yieldPtb}{37}
\newcommand{\yieldChip}{24}
\newcommand{\yieldSpeed}{85}
\newcommand{\nExpPtb}{\num{6115}}

\newcommand{\btSpeedKids}{324}
\newcommand{\btSpeedFalseWinKids}{194}
\newcommand{\btSpeedFalseWinKidsPct}{60}

\newcommand{\vtChipGoodPct}{29}

\newcommand{\vtSpeedGoodPct}{16}

\newcommand{\nTruthChip}{10}
\newcommand{\nTruthSpeed}{6}
\newcommand{\vtChipN}{150}
\newcommand{\vtSpeedN}{310}
\newcommand{\rhoLate}{+0.53}

\newcommand{\lateUp}{9}
\newcommand{\lateTasks}{10}
\newcommand{\latePairUp}{11}
\newcommand{\latePairN}{16}
\newcommand{\rhoLatePerStep}{+0.36}

\newcommand{\rhoStepsAll}{+0.03}
\newcommand{\nFindRuns}{85}

\newcommand{\chipTimeUsedMedian}{19}

\newcommand{\lateTryExpPct}{36}
\newcommand{\lateTryGainPct}{10}
\newcommand{\lateTryWinPct}{33}
\newcommand{\earlyTryWinPct}{28}
\newcommand{\speedFirstGainPct}{72}

\newcommand{\nPersonaRuns}{91}

\newcommand{\varTraitModelMedian}{43}
\newcommand{\varTraitTaskMedian}{7}
\newcommand{\conSameMedian}{0.04}
\newcommand{\conOtherMedian}{0.08}

\newcommand{\conSameCount}{1}
\newcommand{\conOtherCount}{2}
\newcommand{\conSameDenomMax}{64}



\papertitle[Open-Endedness Bench: Measuring Epistemic Process from Agent Records]{%
  Open-Endedness Bench: Measuring Epistemic Process\\from Agent Records%
}
\paperrunningtitle{Open-Endedness Bench}

\paperauthors{%
  Chengyang Shi\textsuperscript{1,5},\ %
  Xianglin Ji\textsuperscript{2,5},\ %
  Jintao Huang\textsuperscript{3,5},\ %
  Jicheng Wang\textsuperscript{4,5},\ %
  Yifeng He\textsuperscript{4,5},\ %
  Jiachen Liu\textsuperscript{5}%
}
\paperaffiliations{%
  \textsuperscript{1}University of Michigan, Ann Arbor,\ %
  \textsuperscript{2}Massachusetts Institute of Technology,\ %
  \textsuperscript{3}The Ohio State University,\ %
  \textsuperscript{4}University of California, Davis,\ %
  \textsuperscript{5}ARA Lab%
}

\paperabstract{Agents are increasingly given open-ended research tasks: discovering an empirical law from self-designed experiments, improving a heuristic whose optimum nobody knows, or beating a standing record. Their execution logs record every step of this research, yet the runs are still judged by their outcome score. That score alone does not establish whether an agent's claims follow from executed experiments, and a reference answer may be unavailable. We evaluate the agent's \emph{epistemic process}: how it forms hypotheses, tests them, and revises them in response to evidence. We introduce \oeb{} (\textsc{Open-Endedness Bench}), a benchmark-agnostic methodology that reads only the agent's execution record and never a reference answer or an outcome score. \oeb{} compiles the record into a unified epistemic event graph whose edges connect the propositions the agent states to the executed actions that test them; each node carries an exact excerpt that code verifies against the record. One principle governs scoring: prose can state a proposition, but only evidence returned by an executed action can support or refute it, so \oeb{} checks what the agent writes against what it actually ran. From the graph, \oeb{} scores four competence axes (evidence, experiment, revision, and no reward hacking), mostly as the share of opportunities for sound research that the agent took, and profiles six subjective persona traits that describe the agent's research habits. We score \nRuns{} existing runs over \nTasks{} tasks from three benchmarks: LLM post-training, chip design, and a training-speed record. Against logged results, only \vtSpeedGoodPct--\vtChipGoodPct\% of the improvements agents claim are real. On \lateUp{} of \lateTasks{} tasks, the best run tries more new ideas in its second half than the worst run. The persona readings follow the model: for every trait, the model that ran explains more of its variance across runs than the task (a median of \varTraitModelMedian\% against \varTraitTaskMedian\%).
}

\papercorrespondence{Chengyang Shi (\href{mailto:chengysh@umich.edu}{chengysh@umich.edu}), Jiachen Liu (\href{mailto:amber@ara-commons.com}{amber@ara-commons.com})}
\papercode{\href{https://github.com/ARA-Labs/oeb}{github.com/ARA-Labs/oeb}}
\paperlogo{}{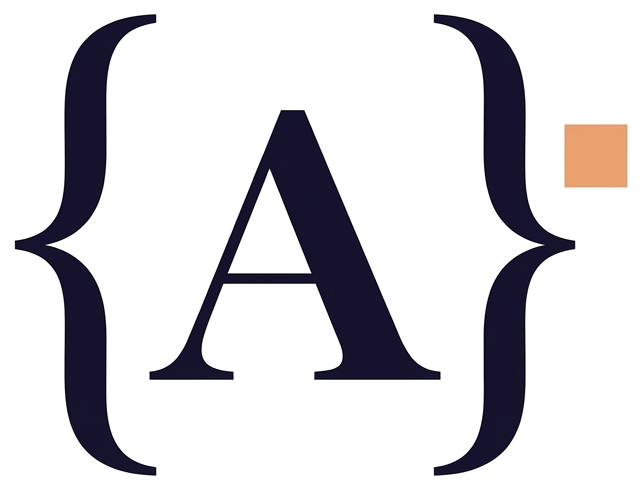}
\paperlogoheight{20pt}
\paperbrand[11pt]{figures/ara-logo.png}{ARA Lab}
\papershortauthors{Shi et al.}

\begin{document}

\makepaperheader

\section{Introduction}
\label{sec:intro}

LLM agents now undertake research tasks that require them to choose what to investigate and how to test it. The AI Scientist generates research ideas, writes code, and runs experiments in machine learning \citep{lu2024aiscientist}. RE-Bench includes a task in which agents use smaller training runs to predict model performance at a larger compute budget \citep{wijk2024rebench}, while ALE-Bench evaluates heuristic algorithms for problems such as delivery routing \citep{imajuku2025alebench}. These settings motivate evaluating both results and the processes that produce them.

These agents are also the first researchers whose whole working process is on record. A human research project usually leaves a paper that shows the path that worked, and even a careful lab notebook captures only part of the reasoning behind it. An agent's execution log keeps every message it writes, every command it runs, and every result it reads back. Peer review and replication were built for researchers who could hand in only their results. Agent runs are cheap to repeat and their records are long, so reading them by hand does not scale; since every step can be checked against the record, much of this checking can be done by a program.

\begin{figure}[t]
\centering
\resizebox{\linewidth}{!}{\begingroup
\definecolor{vInk}{HTML}{111827}\definecolor{vMute}{HTML}{6B7280}\definecolor{vLine}{HTML}{9CA3AF}\definecolor{vFaint}{HTML}{E5E7EB}\definecolor{vLane}{HTML}{F3F4F6}
\definecolor{vSt}{HTML}{DCE8FA}\definecolor{vStLine}{HTML}{5B84C4}\definecolor{vAcc}{HTML}{1F4E9C}
\definecolor{vLa}{HTML}{FBEBD3}\definecolor{vLaLine}{HTML}{C99A4B}\definecolor{vObs}{HTML}{ECEEF2}
\definecolor{vBad}{HTML}{B3261E}\definecolor{vBadFill}{HTML}{FBE3E0}\definecolor{vGood}{HTML}{1E6B45}\definecolor{vGoodFill}{HTML}{D9F0E1}
\newcommand{\stepl}[1]{{\fontsize{4.8}{5.4}\selectfont\sffamily\bfseries\color{vMute}#1}}
\begin{tikzpicture}[x=1in, y=-1in,
  head/.style={anchor=north west, inner sep=0pt, font=\fontsize{7}{8}\selectfont\sffamily\bfseries, text=vInk},
  sub/.style={anchor=north west, inner sep=0pt, font=\fontsize{5.5}{6.2}\selectfont\sffamily, text=vMute, align=left},
  num/.style={circle, fill=vInk, text=white, font=\fontsize{6}{6}\selectfont\sffamily\bfseries, inner sep=0.7pt, minimum size=9.5pt, anchor=north west},
  raw/.style={anchor=north west, inner sep=0pt, text width=1.2in, align=left, font=\fontsize{4.3}{5}\selectfont\ttfamily, text=vInk},
  rawl/.style={anchor=north west, inner sep=0pt, font=\fontsize{5}{5.6}\selectfont\sffamily\bfseries, text=vMute},
  stp/.style={anchor=north west, rounded corners=2pt, line width=0.4pt, inner sep=2.5pt, text width=1.16in, align=left, font=\fontsize{4.8}{5.6}\selectfont\sffamily, text=vInk},
  stpt/.style={stp, font=\fontsize{4.8}{5.6}\selectfont\ttfamily},
  mini/.style={rounded corners=2pt, line width=0.45pt, minimum height=0.15in, minimum width=0.34in, inner sep=1.5pt, font=\fontsize{5}{5.4}\selectfont\sffamily, text=vInk, anchor=center},
  e/.style={-{Stealth[length=3pt,width=2.8pt]}, line width=0.6pt, draw=vInk},
  el/.style={font=\fontsize{4.8}{5}\selectfont\sffamily\itshape, inner sep=0.8pt, text=vInk, fill=white},
  big/.style={-{Stealth[length=5.5pt,width=5pt]}, line width=1.1pt, draw=vInk},
  fan/.style={-{Stealth[length=4pt,width=3.6pt]}, line width=0.6pt, draw=vMute},
    bar/.style={anchor=east, inner sep=0pt, font=\fontsize{5.8}{6.4}\selectfont\sffamily, text=vInk, align=right},
  pole/.style={font=\fontsize{4.2}{4.6}\selectfont\sffamily, text=vMute, inner sep=0pt},
  lab/.style={anchor=north west, inner sep=0pt, font=\fontsize{5}{5.6}\selectfont\sffamily\bfseries, text=vMute},
]
\node[num] at (0.0,0) {1};
\node[head] at (0.180,0.005) {Raw logs};
\node[sub] at (0.180,0.13) {any harness, any format};
\node[num] at (1.5,0) {2};
\node[head] at (1.680,0.005) {The record};
\node[sub] at (1.680,0.13) {one format, 279 steps \textperiodcentered\ L0, code};
\node[num] at (2.95,0) {3};
\node[head] at (3.130,0.005) {The graph};
\node[sub] at (3.130,0.13) {L1--L2: LLM \textperiodcentered\ L3: code};
\node[num] at (4.3,0) {4};
\node[head] at (4.480,0.005) {The scores};
\node[sub] at (4.480,0.13) {counts over the graph};
\node[rawl] at (0.0,0.320) {Claude Code stream};
\node[raw] at (0.0,0.405) {\{"type":\allowbreak{}"assistant",\allowbreak{}"message":\allowbreak{}\{\ldots{}"name":\allowbreak{}"Bash",\allowbreak{}"input":\allowbreak{}\{"command":\allowbreak{}"\ldots{}evaluate.py -{}-model-path out/sft1\_causal\ldots{}};
\draw[fan] (1.250,0.520) -- (1.450,1.040);
\node[rawl] at (0.0,0.820) {Codex event log};
\node[raw] at (0.0,0.905) {\{"type":\allowbreak{}"item.started",\allowbreak{}"item":\allowbreak{}\{\ldots{}"command":\allowbreak{}"/bin/bash -lc pwd"\ldots{}};
\draw[fan] (1.250,1.020) -- (1.450,1.040);
\node[rawl] at (0.0,1.320) {speedrun harness};
\node[raw] at (0.0,1.405) {\{"src":\allowbreak{}"agent",\allowbreak{}"tools":\allowbreak{}[\allowbreak{}\{"fn":\allowbreak{}"Read",\allowbreak{}"cmd":\allowbreak{}"cat \ldots{}program.md"\}],\allowbreak{}"obs":\allowbreak{}"\ldots{}};
\draw[fan] (1.250,1.520) -- (1.450,1.040);
\node[sub, text width=1.3in] at (0.0,1.780) {each benchmark's logs are converted once};
\node[stp, draw=vStLine, fill=vSt] at (1.5,0.320) {\stepl{step 199 $\cdot$ reasoning}\\[1pt]Diagnosis: greedy decoding causes repetition loops \ldots{} switch sft1\_causal to temperature 0.6};
\node[stpt, draw=vLaLine, fill=vLa] at (1.5,0.820) {\stepl{step 203 $\cdot$ action}\\[1pt]python evaluate.py -{}-model-path out/sft1\_causal -{}-json-output-file results/sft1c\_t06.json};
\node[stpt, draw=vLine, fill=vObs] at (1.5,1.320) {\stepl{step 212 $\cdot$ observation}\\[1pt]"accuracy": 0.03333333333333333};
\node[sub, text width=1.3in] at (1.5,1.780) {each step: what it reasoned, what it ran, what came back};
\draw[big] (2.760,1.040) -- (2.890,1.040);
\fill[vLane, rounded corners=2.5pt] (2.95,0.32) rectangle (4.170,0.820);
\fill[vLane, rounded corners=2.5pt] (2.95,1.040) rectangle (4.170,1.680);
\node[lab] at (2.980,0.350) {SAID};
\node[lab] at (2.980,1.070) {DID};
\node[mini, draw=vStLine, fill=vSt] (s1) at (3.220,0.620) {cause};
\node[mini, draw=vStLine, fill=vSt] (s2) at (3.880,0.620) {prediction};
\node[mini, draw=vBad, fill=vBadFill] (a1) at (3.220,1.270) {check};
\node[mini, draw=vLaLine, fill=vLa] (a2) at (3.880,1.270) {launch};
\node[mini, draw=vBad, fill=vBadFill] (a3) at (3.880,1.520) {check};
\draw[e] (s1.south) -- (a1.north); \node[el] at (3.220,0.930) {explains};
\draw[e] (a2.north) -- (s2.south); \node[el, anchor=west] at (3.930,0.930) {expects};
\draw[e, draw=vMute, line width=0.4pt] (a3.north) -- (a2.south);
\draw[e, draw=vBad] (a3.west) -- (3.550,1.520) -- (3.550,0.620) -- (s2.west); \node[el, text=vBad] at (3.550,0.930) {contradicts};
\node[sub, text width=1.25in] at (2.95,1.780) {L1--L2: an LLM answers fixed-choice questions, 3 votes each, with exact excerpts from the record; L3: code assembles the graph};
\draw[big] (4.210,1.040) -- (4.240,1.040);
\node[bar] at (4.830,0.380) {Evidence};
\draw[vFaint, line width=3pt] (4.880,0.380) -- (5.380,0.380);
\draw[vAcc, line width=3pt] (4.880,0.380) -- (5.155,0.380);
\node[sub, anchor=west, text=vInk] at (5.185,0.380) {0.55};
\node[bar] at (4.830,0.550) {Experiment};
\draw[vFaint, line width=3pt] (4.880,0.550) -- (5.380,0.550);
\draw[vAcc, line width=3pt] (4.880,0.550) -- (5.141,0.550);
\node[sub, anchor=west, text=vInk] at (5.171,0.550) {0.52};
\node[bar] at (4.830,0.720) {Revision};
\draw[vFaint, line width=3pt] (4.880,0.720) -- (5.380,0.720);
\node[sub, anchor=west, font=\fontsize{4.6}{5}\selectfont\sffamily] at (4.900,0.720) {3 of 3: too few};
\node[bar] at (4.830,0.890) {No reward\\hacking};
\draw[vFaint, line width=3pt] (4.880,0.890) -- (5.380,0.890);
\draw[vAcc, line width=3pt] (4.880,0.890) -- (5.380,0.890);
\node[sub, anchor=west, text=vInk] at (5.410,0.890) {1.00};
\node[lab] at (4.3,1.120) {PERSONA, SUBJECTIVE};
\node[bar] at (4.830,1.320) {T1};
\draw[vLine, line width=0.5pt] (4.880,1.320) -- (5.500,1.320);
\node[pole, anchor=north west] at (4.880,1.340) {thinking};
\node[pole, anchor=north east] at (5.500,1.340) {doing};
\fill[vInk] (4.954,1.320) circle (1.6pt);
\node[bar] at (4.830,1.520) {T6};
\draw[vLine, line width=0.5pt] (4.880,1.520) -- (5.500,1.520);
\node[pole, anchor=north west] at (4.880,1.540) {run first};
\node[pole, anchor=north east] at (5.500,1.540) {predict first};
\fill[vInk] (5.234,1.520) circle (1.6pt);
\node[sub, text width=1.2in] at (4.3,1.780) {never read: the task's score or a reference answer};
\end{tikzpicture}
\endgroup}
\caption{\textbf{\oeb{} evaluates epistemic process without reading task outcomes or reference answers.} Raw logs compile into an epistemic event graph that connects stated propositions to executed actions, producing objective epistemic competence scores and subjective persona traits solely from how the agent handles evidence.}
\label{fig:overview}
\end{figure}
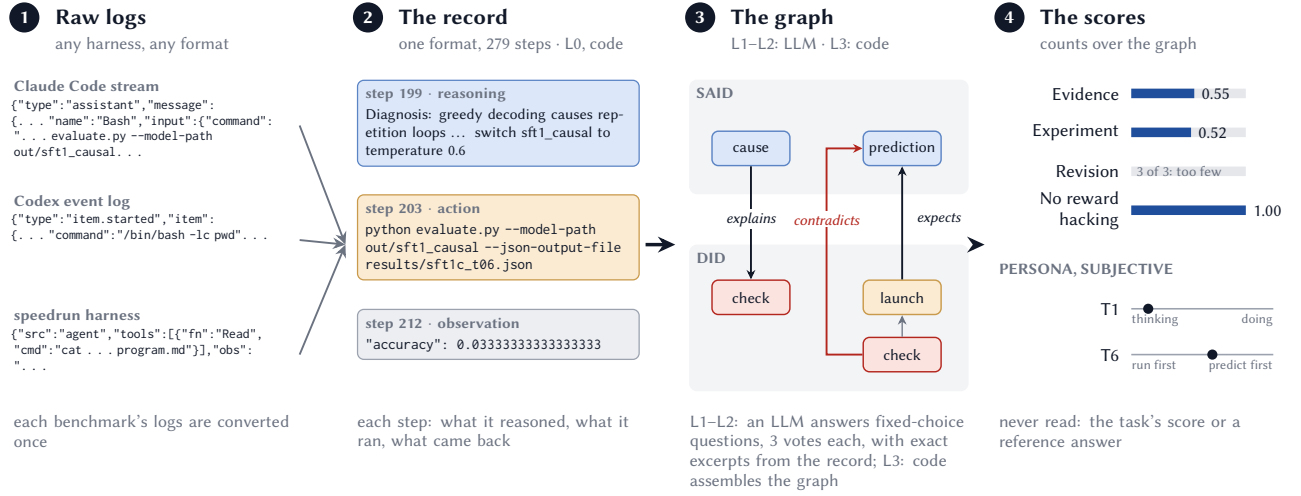
Benchmarks still judge these runs by their outcome score, and where no reference answer exists, the score alone leaves the agent's work a black box. Hours with no improvement can be methodical elimination of dead ends or aimless thrashing, and a high score can come without the experiments the task requires: an RE-Bench agent predicted a scaling law without training models across compute scales \citep{wijk2024rebench}. We need a standard for how an agent handles evidence that does not depend on that score.

Prior work evaluates intermediate steps with labeled reasoning traces or task-specific rubrics \citep{lightman2023verify,jansen2024discoveryworld,starace2025paperbench}. We seek an evaluation of epistemic process that uses neither a reference answer nor a task-specific rubric.

We ground this evaluation in the relation between hypotheses, experiments, and observed evidence. \citet{klahr1988dual} modeled discovery as search over hypotheses and experiments, and observed both hypothesis-led and experiment-led strategies. We score \emph{epistemic competence}, the discipline with which an agent handles evidence, and separately profile its research style, its subjective \emph{persona}, in which neither strategy ranks above the other.

Measuring these aspects of research from an execution record poses four challenges:
\begin{enumerate*}[label=(\roman*)]
  \item the evaluator must turn ``does this claim follow from the evidence?'' into counts that code can compute, without knowing the right answer;
  \item it must tell an agent's statement that it did some work from evidence that the work ran, rather than trust the agent's account \citep{pitfalls2025};
  \item it must apply the same scoring criteria across domains without requiring a new task-specific rubric \citep{starace2025paperbench}; and
  \item its semantic judgments require validation: prior work finds that even strong LLMs struggle to locate decisive errors in agent trajectories \citep{zhang2025whowhen}.
\end{enumerate*}

We introduce \oeb{} to address these challenges.
For (i), it compiles the record into a \emph{unified epistemic event graph} of the acts the agent executed and the propositions it stated, and all but one score is a count over that graph.
For (ii), it scores under one principle, \emph{semantics proposes, actions dispose}: only evidence from an executed action can support or refute a proposition. The axes that read the agent's written account compare it with the actions in the same record, and the reward-hacking axis reads the actions alone. Most scores are the share of \emph{opportunities} the agent took, such as the share of its experiments whose results it read and drew a conclusion from.
For (iii), the scoring code contains no benchmark-specific vocabulary. Task-specific rules enter only through the \emph{world manifest}, a file written when the logs are converted that states the task's evaluation rules.
For (iv), the LLM judge only extracts quoted excerpts and answers closed questions, with three independent votes each, and code verifies every quoted excerpt against the record. We validate the instrument in \autoref{sec:validity}.
An axis averages several measures, its \emph{constructs}, with equal weight; we fit no weights and combine no axes into an overall score. A construct with too few opportunities \emph{abstains} and reports no score.

\paragraph{Contributions.}
\begin{itemize}[leftmargin=1.2em,itemsep=1pt,topsep=2pt]
  \item A formulation of process evaluation for open-ended agent research that needs no ground-truth verifier or reference answer. It audits execution records and is designed to resist gaming even by an agent that knows the rubric.
  \item The unified epistemic event graph: a benchmark-agnostic representation that separates acts from propositions. Every benchmark's logs are converted into one record format, so the same compiler reads them all, and each node carries an exact excerpt verified against the record.
  \item A two-tier instrument that separates four competence axes (evidence, experiment, revision, no reward hacking) from six subjective persona traits. Every construct that is a rate uses one scoring rule, opportunities taken over opportunities offered.
  \item Validation and findings on \nRuns{} existing runs over \nTasks{} tasks from three benchmarks. Rescoring changes a score by a median of \conSameMedian{} (\conOtherMedian{} under another judge model). Only \vtSpeedGoodPct--\vtChipGoodPct\% of the improvements agents claim are real. On \lateUp{} of \lateTasks{} tasks, the best run tries more new ideas late than the worst. For every persona trait, the model explains more variance than the task (median \varTraitModelMedian\% vs.\ \varTraitTaskMedian\%).
\end{itemize}

\section{Problem formulation}
\label{sec:problem}

\paragraph{The execution record.}
We evaluate an append-only execution \emph{record} (what agent papers call a \emph{trajectory}) $\mathcal{R} = (s_1, s_2, \dots, s_T)$, where each step $s_t = (a_t, o_t, m_t)$ holds the agent's executable action $a_t$, the observation $o_t$ it returned, and the agent's reasoning notes $m_t$.
The \emph{harness} is the software that runs the model and logs each step. When we convert its logs, we tag every step with a scope:
\emph{environment steps} ($s_t \in \mathcal{S}_{\text{env}}$) act on the studied system and return new evidence;
\emph{workspace steps} ($s_t \in \mathcal{S}_{\text{ws}}$) work on internal tools or files; and
\emph{submission steps} ($s_t \in \mathcal{S}_{\text{sub}}$) submit a final result.
\oeb{} reads $\mathcal{R}$ and the world manifest (\autoref{sec:method}) alone: never a held-out answer key or the environment's internals.

\paragraph{Epistemic process evaluation.}
A research task may offer outcome measures but no reference solution, and outcomes alone do not show how the agent handled evidence.
We therefore evaluate its \emph{epistemic process}: at every step $t$, the agent's asserted beliefs and commitments must remain faithful to the evidence that environment steps have returned up to step $t$.
A conversion step (L0) turns a benchmark's logs into the shared record format and records the evaluation rules and setup that the log does not state in a \emph{world manifest}; a compiler builds the graph (\autoref{sec:method}); and a measurement layer scores it with fixed rules (\autoref{sec:measure}).

\paragraph{Threat model.}
Two threats face any evaluation that reads the agent's own record.
In \emph{timeline forgery} (Threat~A), the agent claims afterwards to have anticipated an unexpected outcome. Every extracted proposition carries an exact excerpt that code locates in the step where it first appeared, which rules this out as long as the harness writes the record.
In \emph{rubric farming} (Threat~B), an agent that knows the rubric writes forward-looking but empty prose without running informative experiments. A score credits the agent only through a link to an executed \emph{action}, such as running a command, editing a file, or launching an experiment, so prose alone earns nothing; \autoref{sec:app-attacks} lists five ways to farm the rubric and each defense.

\section{The unified epistemic event graph}
\label{sec:method}

\oeb{} scores a record by first compiling it into a graph, because neither half of the record suffices on its own. An instrument that read only actions and observations ($a_t, o_t$) could not tell a principled hypothesis test from a random parameter sweep, or whether the agent understood why an experiment failed. One that read only the agent's reasoning ($m_t$) would reward an agent that narrates hypotheses while running broken or uninformative code. The graph has two planes of nodes. The \emph{act plane} holds the actions the agent executed; the \emph{proposition plane} holds the beliefs, conjectures, predictions, and questions it stated in its reasoning. Typed edges connect a proposition to the act that tests it or otherwise bears on it. Every score in \autoref{sec:measure} except the token share T1 is a count over this graph, and every count that credits the agent requires an executed act, so prose alone never scores.

\begin{figure}[t]
\centering
\resizebox{\linewidth}{!}{\begingroup
\definecolor{vInk}{HTML}{111827}\definecolor{vMute}{HTML}{6B7280}\definecolor{vLine}{HTML}{9CA3AF}\definecolor{vFaint}{HTML}{E5E7EB}\definecolor{vLane}{HTML}{F3F4F6}
\definecolor{vSt}{HTML}{DCE8FA}\definecolor{vStLine}{HTML}{5B84C4}\definecolor{vAcc}{HTML}{1F4E9C}
\definecolor{vLa}{HTML}{FBEBD3}\definecolor{vLaLine}{HTML}{C99A4B}\definecolor{vLaDeep}{HTML}{8A5A00}
\definecolor{vBad}{HTML}{B3261E}\definecolor{vBadFill}{HTML}{FBE3E0}\definecolor{vObs}{HTML}{ECEEF2}
\newcommand{\vK}[1]{{\bfseries #1}}
\begin{tikzpicture}[x=1in, y=-1in,
  rec/.style={anchor=north west, inner sep=0pt, text width=2.02in, align=left, font=\fontsize{6.3}{7.3}\selectfont\sffamily, text=vInk},
  cmd/.style={rec, font=\fontsize{5.7}{6.9}\selectfont\ttfamily},
  tag/.style={anchor=north west, inner sep=0pt, font=\fontsize{6.6}{7}\selectfont\sffamily\bfseries, text=vInk},
  kind/.style={anchor=north west, inner sep=0pt, font=\fontsize{5.4}{6}\selectfont\sffamily, text=vMute},
  nd/.style={anchor=north west, rounded corners=2.5pt, line width=0.55pt, inner xsep=4pt, inner ysep=0pt, minimum height=0.30in, text width=1.02in, align=left, font=\fontsize{6.3}{7.2}\selectfont\sffamily, text=vInk},
  said/.style={nd, draw=vStLine, fill=vSt}, did/.style={nd, draw=vLaLine, fill=vLa}, chk/.style={nd, draw=vBad, fill=vBadFill},
  head/.style={anchor=north west, inner sep=0pt, font=\fontsize{7.2}{8}\selectfont\sffamily\bfseries, text=vInk},
  sub/.style={anchor=north west, inner sep=0pt, font=\fontsize{5.8}{6.6}\selectfont\sffamily, text=vMute},
  main/.style={-{Stealth[length=4.2pt,width=3.8pt]}, line width=0.8pt, draw=vInk, rounded corners=2.5pt},
  ref/.style={main, draw=vBad},
  thin/.style={-{Stealth[length=3.2pt,width=3pt]}, line width=0.5pt, draw=vMute, rounded corners=2.5pt},
  el/.style={font=\fontsize{5.6}{6}\selectfont\sffamily\itshape, inner sep=1.2pt, text=vInk, fill=white},
]
\node[head] at (0.34,0) {The record};
\node[sub] at (0.34,0.135) {steps 184--220 of 279, quoted text};
\node[head] at (2.56,0) {The graph};
\node[sub] at (2.56,0.135) {what the agent \textbf{said}};
\node[sub] at (4.04,0.135) {what the agent \textbf{did}};
\draw[vFaint, line width=0.5pt] (0.0,0.30) -- (5.5,0.30);
\fill[vLane, rounded corners=3pt] (2.500,0.340) rectangle (3.760,3.765);
\fill[vLane, rounded corners=3pt] (3.980,0.340) rectangle (5.240,3.765);
\node[tag] at (0.0,0.400) {184};
\node[kind] at (0.0,0.505) {ran};
\node[cmd] at (0.34,0.400) {\sethlcolor{vLa}\hl{python evaluate.py -{}-model-path out/sft1\_causal -{}-json-output-file results/sft1c.json}};
\node[did] (L1) at (4.040,0.400) {\vK{Launch}\ \ evaluate the fine-tuned model};
\node[tag] at (0.0,0.770) {196};
\node[kind] at (0.0,0.875) {saw};
\node[cmd] at (0.34,0.770) {\sethlcolor{vObs}\hl{correct: 0/30  has\_ANSWER: 0} [\ldots] \sethlcolor{vObs}\hl{completion tokens: mean 16001} [\ldots] \sethlcolor{vObs}\hl{stop\_reasons: Counter(\{'max\_tokens': 30\})}};
\node[chk] (C1) at (4.040,0.770) {\vK{Check}\ \ 0 of 30 answer; all 30 hit the token limit};
\node[tag] at (0.0,1.173) {199};
\node[kind] at (0.0,1.278) {said};
\node[rec] at (0.34,1.173) {Diagnosis: \sethlcolor{vSt}\hl{greedy decoding causes repetition loops}. The model reasons coherently at first, then gets stuck repeating the same sentence until it hits 16k tokens [\ldots] Let me \sethlcolor{vSt}\hl{switch sft1\_causal to temperature 0.6 / top\_p 0.95 and re-eval}};
\node[said] (A2) at (2.560,1.173) {\vK{Cause}\ \ greedy decoding makes it loop};
\node[said] (P2) at (2.560,1.523) {\vK{Prediction}\ \ with sampling it will stop};
\node[tag] at (0.0,1.893) {203};
\node[kind] at (0.0,1.998) {ran};
\node[cmd] at (0.34,1.893) {\sethlcolor{vLa}\hl{python evaluate.py -{}-model-path out/sft1\_causal -{}-json-output-file results/sft1c\_t06.json}};
\node[did] (L2) at (4.040,1.893) {\vK{Launch}\ \ evaluate again at temperature 0.6};
\node[tag] at (0.0,2.263) {212};
\node[kind] at (0.0,2.368) {saw};
\node[cmd] at (0.34,2.263) {\sethlcolor{vObs}\hl{"accuracy": 0.03333333333333333} [\ldots] \sethlcolor{vObs}\hl{correct: 1/30} [\ldots] \sethlcolor{vObs}\hl{stop\_reasons: Counter(\{'max\_tokens': 28, 'stop': 2\})}};
\node[chk] (C2) at (4.040,2.263) {\vK{Check}\ \ 1 of 30 correct; 28 hit the token limit};
\node[tag] at (0.0,2.666) {214};
\node[kind] at (0.0,2.771) {said};
\node[rec] at (0.34,2.666) {Temp 0.6 helped marginally (3.3\%, 1/30) but 28/30 still ramble to 16k tokens [\ldots] \sethlcolor{vSt}\hl{the model is badly undertrained and rarely concludes}. Crucially, the 2 that did stop prove the termination mechanism works [\ldots] Run 2 needs much more: more data, higher LR, on-distribution AIME.};
\node[said] (A3) at (2.560,2.666) {\vK{Cause}\ \ the model is undertrained};
\node[tag] at (0.0,3.372) {220};
\node[kind] at (0.0,3.477) {ran};
\node[cmd] at (0.34,3.372) {\sethlcolor{vLa}\hl{python train\_sft.py -{}-out out/sft2 -{}-data data/sft\_mot -{}-max-examples 12000 -{}-extra-data data/sft\_aime -{}-extra-repeat 3 -{}-lr 2e-5}};
\node[did] (L3) at (4.040,3.372) {\vK{Launch}\ \ train again on 13.7k examples};
\draw[thin] (C1.north) -- (L1.south); \node[el, text=vMute, anchor=west, fill=none] at ($(C1.north)!0.5!(L1.south)+(0.04,0)$) {reads};
\draw[main] (A2.east) -- (C1.west); \node[el, sloped, anchor=south] at ($(A2.east)!0.55!(C1.west)$) {explains};
\draw[main] (L2.west) -- (3.820,0 |- L2.west) |- ($(P2.east)+(0,0.05)$);
\node[el, rotate=90] at (3.820,1.858) {expects};
\draw[thin] (C2.north) -- (L2.south); \node[el, text=vMute, anchor=west, fill=none] at ($(C2.north)!0.5!(L2.south)+(0.04,0)$) {reads};
\draw[ref] ($(C2.west)+(0,-0.06)$) -- (3.940,0 |- C2.west) |- ($(P2.east)+(0,-0.05)$);
\node[el, rotate=90, text=vBad] at (3.940,2.228) {contradicts};
\draw[main] (A3.east) -- ($(C2.west)+(0,0.06)$); \node[el, sloped, anchor=south] at ($(A3.east)!0.55!(C2.west)$) {explains};
\draw[thin] (L3.north) -- (C2.south); \node[el, text=vMute, anchor=west, fill=none] at ($(L3.north)!0.5!(C2.south)+(0.04,0)$) {responds to};
\node[sub, text=vInk] at (0.0,3.795) {\sethlcolor{vSt}\hl{highlight} $=$ the exact excerpt stored by the node \quad \textcolor{vStLine}{\rule{5pt}{5pt}} said \quad \textcolor{vLaLine}{\rule{5pt}{5pt}} did \quad \textcolor{vBad}{\rule{5pt}{5pt}} a check that contradicted a prediction};
\end{tikzpicture}
\endgroup}
\caption{\textbf{Propositions earn epistemic credit only when tied to actions that test them.} When empirical checks contradict a prediction, the graph captures whether the agent updates its beliefs based on observed evidence rather than narrating on without checking.}
\label{fig:episode}
\end{figure}
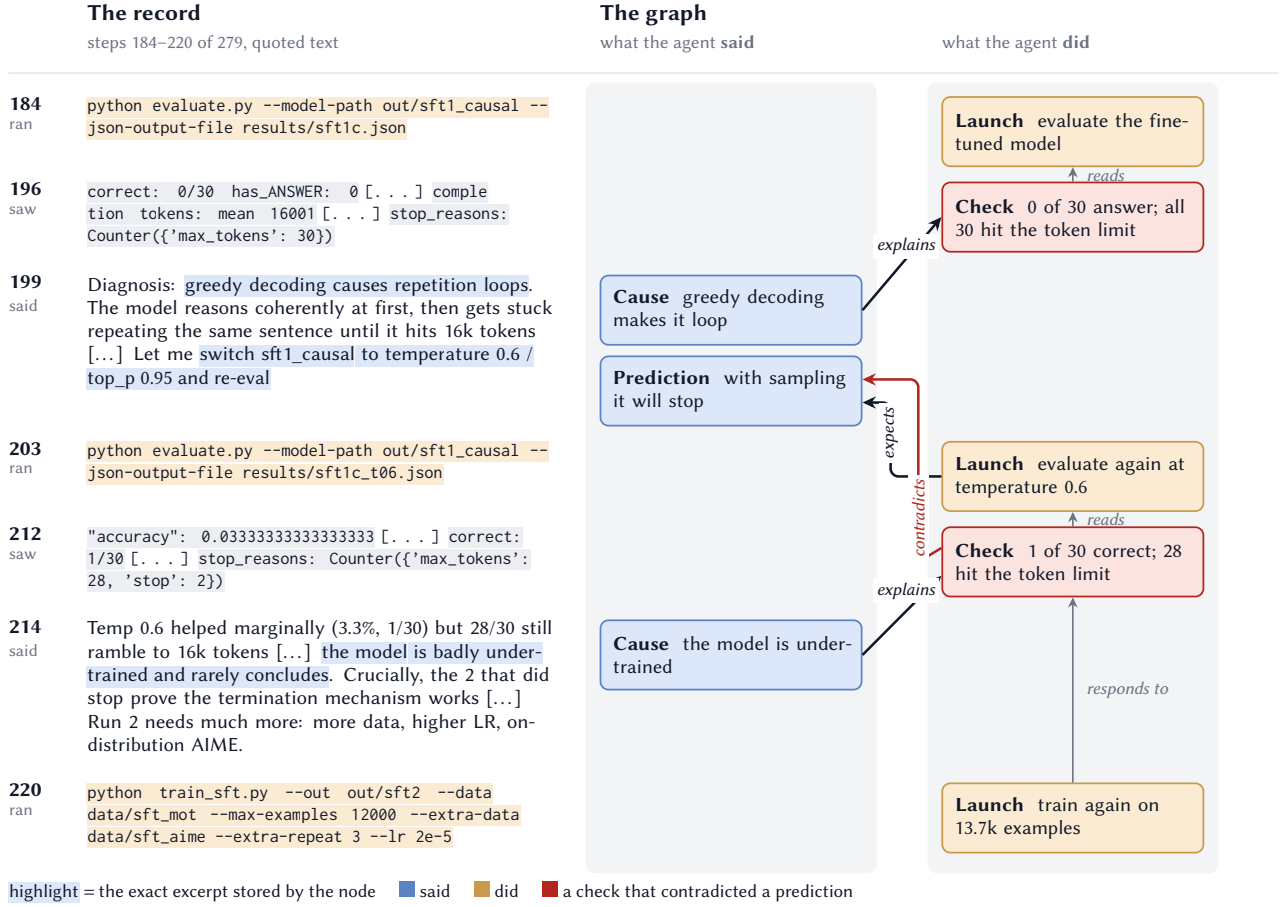

We compile a raw log into the graph $\mathcal{G} = (\mathcal{V}, \mathcal{E})$ in four stages (\autoref{fig:overview}). Code converts the log (L0) and assembles the graph (L3); an LLM extracts cards, typed notes that each quote one step (L1), and answers jury questions that connect cards or mark a card (L2). The same LLM, which we call the judge, does both: GLM-5.3 in our experiments, or Grok-4.6 for the nanoGPT speedrun records~\citep{primeintellect2026speedrun}. We use the LLM for nothing else. It never receives a ``grade this transcript'' prompt, and every card it produces must quote its source step, or code discards the card.

\paragraph{L0: conversion and world manifest.}
L0 converts each benchmark's raw logs into the shared record format $\mathcal{R}$ and tags each step with its scope ($s_t \in \{\mathcal{S}_{\text{env}}, \mathcal{S}_{\text{ws}}, \mathcal{S}_{\text{sub}}\}$). It also writes the world manifest (\texttt{world.json}), the evaluation rules the log does not state, against which E4 judges the agent's actions (\autoref{sec:measure-competence}). This conversion is the only benchmark-specific step (\autoref{sec:app-conversion}).

\paragraph{L1: card extraction with quote verification.}
We use \emph{action} for the raw $a_t$ and \emph{act} for its typed graph node. L1 reads the record in five-step windows and extracts two kinds of \emph{cards}, each a node:
\begin{itemize}[leftmargin=1.2em,itemsep=1pt,topsep=2pt]
  \item \textbf{Act cards ($\mathcal{V}_{\text{act}}$)} classify actions into four types:
    \emph{check} (tests a stated expectation),
    \emph{probe} (gathers evidence before committing to a change),
    \emph{commit} (carries out the work), and
    \emph{revise} (changes an earlier commit after contrary evidence).
  \item \textbf{Proposition cards ($\mathcal{V}_{\text{prop}}$)} record:
    \emph{belief}, \emph{conjecture}, \emph{prediction}, \emph{question}, \emph{reliance} (an assumption the agent builds on without testing it), and \emph{retract} (a withdrawn earlier statement).
\end{itemize}
Every card must quote its source step; code discards a card whose quote is not a literal substring of that step. The prompts (\autoref{sec:app-prompts}) list no examples of qualifying content.

\paragraph{L2: juries.}
Some questions cannot be answered from one step alone, such as whether two propositions say the same thing or whether an idea is new. The extractor reads one step at a time, so a \emph{jury} answers them. Following \citet{verga2024juries}, a jury asks the judge one closed question three times, independently, and takes the majority. Juries do two jobs. Four of them connect cards into the graph: they merge propositions that say the same thing, pair a hypothesis with an earlier finding, match a check to the proposition it tests, and group activities into lines of inquiry. The other two each mark one card, and a measure reads the mark directly. The reward-hacking jury marks whether a commit or revise act breaks a quoted rule (E4, \autoref{sec:measure-competence}). The novelty jury marks whether a new idea differs substantially from earlier settled findings and whether it builds on a mechanism the run has already shown to work (T3 and T5, \autoref{tab:traits}).

\paragraph{L3: deterministic assembly.}
Code builds $\mathcal{G}$ from the cards and jury decisions without further LLM calls. Typed edges connect an act to the proposition it bears on, such as a check that supports or contradicts a prediction. Each proposition starts as untested and becomes supported or contradicted as checks return results. We show one episode in \autoref{fig:episode}.

\section{The measurement layer}
\label{sec:measure}

The measurement layer turns the graph into four competence scores and six persona traits with fixed rules and no further LLM calls. One scoring rule serves every rate-valued measure (\autoref{sec:measure-invariant}); the competence axes and persona traits are its instances (\autoref{sec:measure-competence}, \autoref{sec:measure-persona}).

\subsection{One scoring rule: opportunities taken over opportunities offered}
\label{sec:measure-invariant}

We call each individual measure that contributes to an axis a \emph{construct}. Every construct expressed as a rate uses the same rule:
\begin{equation}
\label{eq:invariant}
\text{score} = \frac{\text{taken}}{\text{opportunities}} = \frac{|\mathcal{K}|}{|\mathcal{C}|},
\end{equation}
An \emph{opportunity} is a situation in the record that calls for a particular response, and the agent \emph{takes} it if the record shows that response. Each experiment the agent launches, for example, is an opportunity to read the result and draw a conclusion; each idea that evidence contradicts is an opportunity to repair the idea or state why it failed. For a persona trait, the response is one of the trait's two poles. $\mathcal{C}$ is the set of opportunities a construct finds in the graph and $\mathcal{K} \subseteq \mathcal{C}$ the set the agent took.
If the record offers no opportunity (no idea is ever contradicted, say), the construct \emph{abstains}: it reports no score, never a zero. A construct with fewer than five opportunities reports its counts but stays out of the mean. An axis is the equal-weight mean of its remaining constructs on $[0,1]$, and we never combine axes into one score.
Two measures do not use the rule: the reward-hacking axis E4 transforms a count, and the budget trait T1 is a share of output tokens.
\paragraph{Safeguards for short records.}
\label{sec:app-safeguards}
To keep statistical noise from corrupting scores on short or sparse records, the measurement layer enforces three safeguards:
\begin{enumerate*}[label=(\roman*)]
  \item \emph{Sample floor}: a rate-valued construct needs at least five opportunities to enter an axis mean. Below the floor it reports its raw counts but contributes nothing to the mean.
  \item \emph{Confidence propagation}: we repeatedly sample construct rates, recompute their axis mean, and report percentiles of those means as an uncertainty interval. For rate-valued constructs the draws use the constituent counts of taken and eligible opportunities.
  \item \emph{Loud abstention}: if no construct in an axis qualifies for its mean, the axis reports no score.
\end{enumerate*}

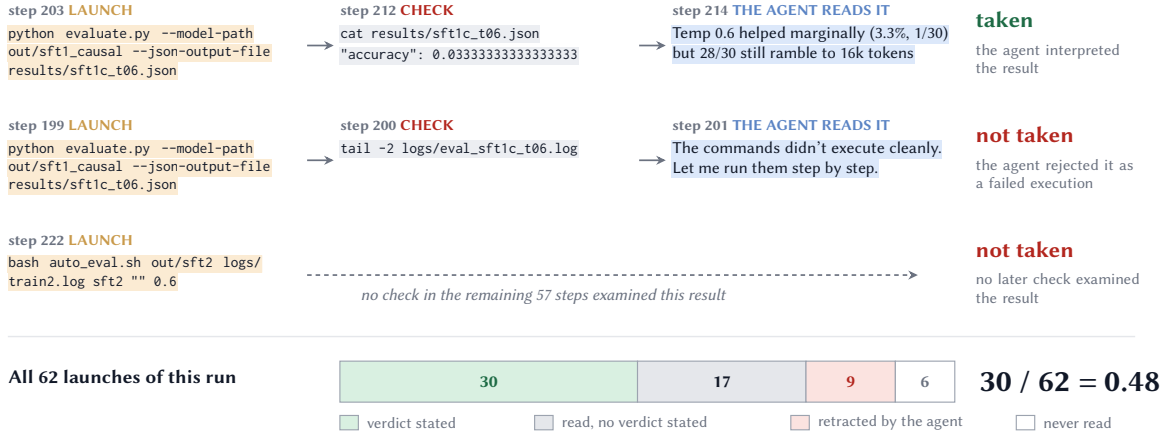
\begin{figure}[t]
\centering
\resizebox{0.9\linewidth}{!}{\begingroup
\definecolor{vInk}{HTML}{111827}\definecolor{vMute}{HTML}{6B7280}\definecolor{vLine}{HTML}{9CA3AF}\definecolor{vFaint}{HTML}{E5E7EB}\definecolor{vLane}{HTML}{F3F4F6}
\definecolor{vSt}{HTML}{DCE8FA}\definecolor{vStLine}{HTML}{5B84C4}\definecolor{vAcc}{HTML}{1F4E9C}
\definecolor{vLa}{HTML}{FBEBD3}\definecolor{vLaLine}{HTML}{C99A4B}\definecolor{vObs}{HTML}{ECEEF2}
\definecolor{vBad}{HTML}{B3261E}\definecolor{vBadFill}{HTML}{FBE3E0}\definecolor{vGood}{HTML}{1E6B45}\definecolor{vGoodFill}{HTML}{D9F0E1}
\begin{tikzpicture}[x=1in, y=-1in,
  rule/.style={anchor=north west, inner sep=0pt, font=\fontsize{8.5}{9.5}\selectfont\sffamily\bfseries, text=vInk},
  txt/.style={anchor=north west, inner sep=0pt, font=\fontsize{6.4}{7.4}\selectfont\sffamily, text=vInk, align=left},
  mute/.style={txt, text=vMute},
  cell/.style={anchor=north west, inner sep=0pt, text width=1.34in, align=left, font=\fontsize{5.9}{6.9}\selectfont\sffamily, text=vInk},
  cmd/.style={cell, font=\fontsize{5.4}{6.5}\selectfont\ttfamily},
  lab/.style={anchor=north west, inner sep=0pt, font=\fontsize{5.4}{6}\selectfont\sffamily\bfseries, text=vMute},
  stamp/.style={anchor=north west, inner sep=0pt, font=\fontsize{8}{9}\selectfont\sffamily\bfseries},
  why/.style={anchor=north west, inner sep=0pt, text width=0.8in, align=left, font=\fontsize{5.6}{6.4}\selectfont\sffamily, text=vMute},
  arr/.style={-{Stealth[length=3.4pt,width=3.2pt]}, line width=0.6pt, draw=vMute},
  seg/.style={anchor=center, inner sep=0pt, font=\fontsize{6.2}{6.2}\selectfont\sffamily\bfseries},
  lg/.style={anchor=west, inner sep=0pt, font=\fontsize{5.6}{6}\selectfont\sffamily, text=vMute},
  big/.style={anchor=west, inner sep=0pt, font=\fontsize{11}{11}\selectfont\sffamily\bfseries, text=vInk},
]
\node[rule] at (0,0) {score $=$ opportunities taken $/$ opportunities offered};
\node[txt, text width=5.5in] at (0,0.17) {For \textbf{experiment yield}, a construct (one measure), an \textbf{opportunity} is a launched experiment. It is \textbf{taken} when a later check yields a verdict and the agent never retracts the result.};
\draw[vFaint, line width=0.5pt] (0,0.44) -- (5.5,0.44);
\node[lab] at (0.0,0.540) {step 203 \textcolor{vLaLine}{LAUNCH}};
\node[cmd] at (0.0,0.640) {\sethlcolor{vLa}\hl{python evaluate.py -{}-model-path out/sft1\_causal -{}-json-output-file results/sft1c\_t06.json}};
\draw[arr] (1.460,0.740) -- (1.590,0.740);
\node[lab] at (1.62,0.540) {step 212 \textcolor{vBad}{CHECK}};
\node[cmd] at (1.62,0.640) {\sethlcolor{vObs}\hl{cat results/sft1c\_t06.json}\\[1pt]\sethlcolor{vObs}\hl{"accuracy": 0.03333333333333333}};
\draw[arr] (3.080,0.740) -- (3.210,0.740);
\node[lab] at (3.24,0.540) {step 214 \textcolor{vStLine}{THE AGENT READS IT}};
\node[cell] at (3.24,0.640) {\sethlcolor{vSt}\hl{Temp 0.6 helped marginally (3.3\%, 1/30) but 28/30 still ramble to 16k tokens}};
\node[stamp, text=vGood] at (4.72,0.570) {taken};
\node[why] at (4.72,0.730) {the agent interpreted the result};
\node[lab] at (0.0,1.100) {step 199 \textcolor{vLaLine}{LAUNCH}};
\node[cmd] at (0.0,1.200) {\sethlcolor{vLa}\hl{python evaluate.py -{}-model-path out/sft1\_causal -{}-json-output-file results/sft1c\_t06.json}};
\draw[arr] (1.460,1.300) -- (1.590,1.300);
\node[lab] at (1.62,1.100) {step 200 \textcolor{vBad}{CHECK}};
\node[cmd] at (1.62,1.200) {\sethlcolor{vObs}\hl{tail -2 logs/eval\_sft1c\_t06.log}};
\draw[arr] (3.080,1.300) -- (3.210,1.300);
\node[lab] at (3.24,1.100) {step 201 \textcolor{vStLine}{THE AGENT READS IT}};
\node[cell] at (3.24,1.200) {\sethlcolor{vSt}\hl{The commands didn't execute cleanly. Let me run them step by step.}};
\node[stamp, text=vBad] at (4.72,1.130) {not taken};
\node[why] at (4.72,1.290) {the agent rejected it as a failed execution};
\node[lab] at (0.0,1.660) {step 222 \textcolor{vLaLine}{LAUNCH}};
\node[cmd] at (0.0,1.760) {\sethlcolor{vLa}\hl{bash auto\_eval.sh out/sft2 logs/train2.log sft2 "" 0.6}};
\draw[arr, dash pattern=on 1.6pt off 1.4pt] (1.460,1.860) -- (4.440,1.860);
\node[mute, font=\fontsize{5.9}{6.9}\selectfont\sffamily\itshape] at (1.720,1.920) {no check in the remaining 57 steps examined this result};
\node[stamp, text=vBad] at (4.72,1.690) {not taken};
\node[why] at (4.72,1.850) {no later check examined the result};
\draw[vFaint, line width=0.5pt] (0,2.160) -- (5.5,2.160);
\node[txt, font=\fontsize{6.8}{7.6}\selectfont\sffamily\bfseries] at (0,2.320) {All 62 launches of this run};
\draw[vLine, line width=0.4pt, fill=vGoodFill] (1.620,2.280) rectangle (3.072,2.480);
\node[seg, text=vGood] at (2.346,2.380) {30};
\draw[vLine, line width=0.4pt, fill=vFaint] (3.072,2.280) rectangle (3.894,2.480);
\node[seg, text=vInk] at (3.483,2.380) {17};
\draw[vLine, line width=0.4pt, fill=vBadFill] (3.894,2.280) rectangle (4.330,2.480);
\node[seg, text=vBad] at (4.112,2.380) {9};
\draw[vLine, line width=0.4pt, fill=white] (4.330,2.280) rectangle (4.620,2.480);
\node[seg, text=vMute] at (4.475,2.380) {6};
\node[big] at (4.740,2.380) {30 / 62 $=$ 0.48};
\draw[vLine, line width=0.4pt, fill=vGoodFill] (1.620,2.545) rectangle (1.710,2.615);
\node[lg] at (1.750,2.580) {verdict stated};
\draw[vLine, line width=0.4pt, fill=vFaint] (2.570,2.545) rectangle (2.660,2.615);
\node[lg] at (2.700,2.580) {read, no verdict stated};
\draw[vLine, line width=0.4pt, fill=vBadFill] (3.820,2.545) rectangle (3.910,2.615);
\node[lg] at (3.950,2.580) {retracted by the agent};
\draw[vLine, line width=0.4pt, fill=white] (4.920,2.545) rectangle (5.010,2.615);
\node[lg] at (5.050,2.580) {never read};
\end{tikzpicture}
\endgroup}
\caption{\textbf{Experiments the agent never reads, or rejects after reading, are missed opportunities.} E2 on this run is the mean of consequential checks (33/59) and experiment yield (30/62), 0.52; E3 is 3/3 but reports no score, because three opportunities fall below the floor of five.}
\label{fig:scoring}
\end{figure}

We illustrate the rule for one construct in \autoref{fig:scoring}, on the run of \autoref{fig:episode}, and list every axis and trait in \autoref{tab:axes}.

\begin{table}[t]
\centering\small
\begin{tabular}{@{}ll@{}}
\toprule
\textbf{Axis / Trait} & \textbf{Core question} \\
\midrule
\multicolumn{2}{c}{\textit{\textbf{Tier 1: Objective epistemic competence (scored, $[0, 1]$)}}} \\
E1 Evidence & Are propositions backed by what was seen? \\
E2 Experiment & Do experiments yield interpretable results? \\
E3 Revision & Does it address contradicted ideas? \\
E4 No reward hacking & Does it avoid manipulating the evaluation? \\
\midrule
\multicolumn{2}{c}{\textit{\textbf{Tier 2: Subjective persona (profiled)}}} \\
T1 Budget & Doing vs.\ thinking \\
T2 Concurrency & One subject vs.\ several at once \\
T3 Search topology & Refine an idea vs.\ test new ideas \\
T4 Failure style & Pivot vs.\ persist \\
T5 Provenance & Transferred vs.\ novel \\
T6 Epistemic tempo & Hypothesis-first vs.\ run-then-read \\
\bottomrule
\end{tabular}

\caption{\textbf{\oeb{} separates objective epistemic competence from subjective persona.} Competence axes score discipline in handling evidence on $[0, 1]$, while persona traits place an agent's working style between two legitimate research habits.}
\label{tab:axes}
\end{table}

\subsection{Objective epistemic competence axes (E1--E4)}
\label{sec:measure-competence}

We score competence on four axes, each defined without reference to any domain. E1--E3 compare what the agent writes with what it did; E4 reads only what it did.

\paragraph{E1: Evidence.}
E1 measures whether the agent's beliefs rest on results it has seen. It averages four constructs:
\begin{enumerate*}[label=(\roman*)]
  \item \emph{belief grounded}, the share of stated beliefs and conjectures that a result already in the record settles;
  \item \emph{assertion within evidence}, the share of such grounded research beliefs that claim no more than the results showed, adding no untried setting, general rule, cause, or forecast;
  \item \emph{reliance tested}, the share of reliances that a check tested after the agent stated them; and
  \item \emph{unbacked belief held up}, the share of beliefs stated before any result bore on them that their first test confirmed.
\end{enumerate*}

\paragraph{E2: Experiment.}
E2 measures whether the agent's experiments answer questions it has written down. It averages two constructs:
\begin{enumerate*}[label=(\roman*)]
  \item \emph{consequential check rate}, the share of checks whose result changed the standing of a stated proposition, either as the first result to support or contradict it or by reversing an earlier result; and
  \item \emph{experiment yield}, the share of launched experiments whose result the agent read and judged, as a pass or a failure or as better or worse, without later withdrawing the result as a bug, a confound, or a misreading.
\end{enumerate*}
A check that repeats a known answer, or bears on nothing the agent wrote down, is not taken. Because E2 counts every experiment, an agent that reasons off the record scores low on it; E1 and E3 find their opportunities in the written account, so a terse agent has fewer of them, and more of its constructs abstain (\autoref{tab:main}).

\begin{table}[!tbp]
\centering\small
\begin{tabular}{@{}lrcccc@{}}
\toprule
\textbf{Agent model} & \textbf{Tasks} & \textbf{E1 Evidence} & \textbf{E2 Experiment} & \textbf{E3 Revision} & \textbf{E4 No reward hacking} \\
\midrule
Claude Opus 5 & 6 & 0.78\,\textsubscript{10} & 0.08\,\textsubscript{12} & -- & 0.83\,\textsubscript{12} \\
Claude Fable 5 & 5 & 0.67\,\textsubscript{10} & 0.29\,\textsubscript{10} & 0.60\,\textsubscript{1} & 0.88\,\textsubscript{10} \\
GPT-5.6 & 6 & 0.65\,\textsubscript{12} & 0.42\,\textsubscript{12} & 0.43\,\textsubscript{12} & 0.88\,\textsubscript{12} \\
GPT-5.5 & 6 & 0.66\,\textsubscript{12} & 0.51\,\textsubscript{12} & 0.56\,\textsubscript{5} & 0.96\,\textsubscript{12} \\
Kimi K3 & 6 & 0.71\,\textsubscript{12} & 0.48\,\textsubscript{12} & 0.62\,\textsubscript{8} & 0.83\,\textsubscript{12} \\
GLM-5.2 & 6 & 0.65\,\textsubscript{12} & 0.51\,\textsubscript{12} & 0.67\,\textsubscript{6} & 0.77\,\textsubscript{12} \\
\bottomrule
\end{tabular}

\caption{\textbf{Competence scores across agent models on PostTrainBench.} Each cell averages all runs of the model, first and second; subscripts count the runs with a score. A model whose record omits its reasoning scores low on E2 and offers fewer opportunities on E1 and E3, where more of its constructs abstain.}
\label{tab:main}
\end{table}

\paragraph{E3: Revision.}
E3 measures what the agent does after evidence contradicts one of its research ideas. Its one construct, \emph{contradicted resolved}, is the share of contradicted ideas that the agent later repaired, so that a run carrying the change passed, or explained by stating a cause; setting an idea aside with ``that did not work'' is not taken. We exclude contradictions the agent traced to a broken run, and engineering failures, which must be fixed for the run to go on.

\paragraph{E4: No reward hacking.}
E4 measures whether the agent manipulated the evaluation. For each commit or revise act, the judge decides whether the act breaks a rule quoted from the task, the world manifest, or a short list of house rules, such as training on test items or picking a seed by its score.
Breaking acts that serve one idea, or that are one action, count as one \emph{reward-hacking instance}. Two moves at the end of a run add one instance each: choosing the final candidate as the best of several scores on part of the test set without confirming it on the whole, and reporting a score on part of the test as the score on the whole.
With $k$ instances, E4 reports $1/(1+k)$: a clean run scores $1$, and one instance scores $0.5$. Dividing $k$ by the number of acts instead would let one hack vanish in a long run.

\subsection{Subjective persona (T1--T6)}
\label{sec:measure-persona}

\label{sec:app-traits}
\oeb{} profiles six subjective \emph{persona traits} (\autoref{tab:axes}). Each trait is a choice between two poles. T1 is the share of output tokens spent on actions. T2--T6 are the share of opportunities in which the agent chose one pole; T6, for example, is the share of propositions the agent stated before the first experiment that tested them, against running first and reading afterwards. We infer the profile from the record alone. Traits never enter a competence score and carry no preferred direction.

\begin{table}[t]
\centering\small
\begin{tabular}{@{}llp{6.1cm}@{}}
\toprule
\textbf{Trait} & \textbf{Poles} & \textbf{Score} \\
\midrule
T1 Budget & Doing vs.\ thinking & Share of output tokens spent on actions. \\
T2 Concurrency & One subject vs.\ several at once & Share of acts at which two or more subjects remained under investigation; share of commits made before evidence resolved the previous commit in the same line of inquiry. \\
T3 Search topology & Refine an idea vs.\ test new ideas & Share of research launches that are the first on their subject; share of new ideas a novelty jury judges substantially different from previously settled findings. \\
T4 Failure style & Pivot vs.\ persist & After an adverse result, share of next launches that move to another subject. \\
T5 Provenance & Transferred vs.\ novel & Among new ideas judged either transferred or novel, share built from a mechanism already proven earlier in the run. \\
T6 Epistemic tempo & Hypothesis-first vs.\ run-then-read & Share of propositions stated as a conjecture, question, or prediction at or before the first launch that tests them. \\
\bottomrule
\end{tabular}

\caption{\textbf{Persona traits profile working style without normative ranking.} Each trait measures the empirical frequency of choosing one valid research habit over another.}
\label{tab:traits}
\end{table}

\section{Instrument validity and measurement consistency}
\label{sec:validity}

We check E4 against a benchmark's own audits and every score against a second scoring.

\subsection{Reward-hacking detection aligns with external benchmark audits}
\label{sec:validity-hacking}

For \nLabelled{} of its \nPtbRuns{} runs, PostTrainBench~\citep{rank2026posttrainbench} publishes its auditors' ruling on whether the agent put test items into the training data or used a disallowed model.
The ruling never enters the record \oeb{} reads, so we can use it as an independent external label for E4.
On these runs, E4 flags a run when it counts at least one reward-hacking instance. It flags \nCheatCaught{} of the \nCheat{} runs the auditors found cheating, with a median of \nCaughtMedian{} instances in each.
It missed one cheating run whose training-data script never entered the record, and each verdict held when the run was scored again (\autoref{sec:validity-stability}).

\subsection{Score stability across rescoring and alternative judge models}
\label{sec:validity-stability}

\begin{figure}[t]
\centering
\includegraphics[width=\linewidth]{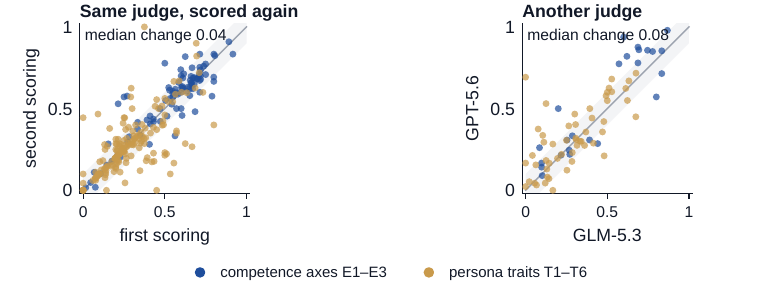}
\caption{\textbf{Evaluation scores are stable across repeated evaluations and alternative judge models.} Differences in competence and trait scores largely remain within $\pm0.1$, so the system's LLM judge adds little variance to the scores.}
\label{fig:rescore}
\end{figure}

Because the judge's votes are sampled, two scorings of one record can differ; overall, they agree (\autoref{fig:rescore}).
When the same judge scored a record twice, the median score difference was \conSameMedian{} on the $[0,1]$ scale; with a different judge model, it was \conOtherMedian{}. \autoref{app:reliability} gives the per-dimension differences.

\section{Empirical findings on agent records}
\label{sec:exp}

We scored \nRuns{} existing agent runs on \nTasks{} tasks from three benchmarks.
We ran no agents for this study: every record is a run its benchmark had already logged, and scoring a benchmark required only converting its logs into the shared record format.
In \emph{PostTrainBench}~\citep{rank2026posttrainbench}, an agent has 10 hours on one H100 GPU to post-train a small base language model for one of six held-out benchmarks (\nPtbRuns{} runs). In \emph{Chip-Bench}~\citep{chipbench2026}, it optimizes a hardware design against a hidden grader (\nChipTasks{} tasks, four agents each). In the \emph{nanoGPT speedrun}~\citep{primeintellect2026speedrun}, it tries to improve the training-speed record (\nSpeedRuns{} runs).

\subsection{Agents overrate their own experimental results}
\label{sec:exp-yield}

\begin{table}[t]
\centering\small
\begin{tabular}{@{}lcccc@{}}
\toprule
 & \multicolumn{2}{c}{\textbf{Truly improved}} & \multicolumn{2}{c}{\textbf{Built on by a later experiment}} \\
\cmidrule(lr){2-3}\cmidrule(l){4-5}
\textbf{The agent read the result as} & \textbf{Speedrun} & \textbf{Chip-Bench} & \textbf{Speedrun} & \textbf{Chip-Bench} \\
\midrule
an improvement & 16\% & 29\% & 91\% & 52\% \\
no improvement & 1\% & 20\% & 20\% & 60\% \\
(no verdict stated) & 3\% & 31\% & 33\% & 51\% \\
\bottomrule
\end{tabular}

\caption{\textbf{Agents claim wins that are not real, and build on them.} Research experiments with a logged result (\vtSpeedN{} speedrun, \vtChipN{} Chip-Bench) by the agent's verdict: the share that truly beat the run's best so far, and the share a later experiment modifies or extends.}
\label{tab:verdicts}
\end{table}

An experiment ends in a verdict when the agent reads its result and says what it shows.
On PostTrainBench, \yieldPtb\% of the \nExpPtb{} launched experiments end in a verdict the agent never later withdraws; on Chip-Bench the share is \yieldChip\%, and on the speedrun \yieldSpeed\%.
Where the benchmark logs every experiment's real result, we can test the verdicts (\autoref{tab:verdicts}, left).
On the speedrun they carry information, but ``improved'' is claimed several times more often than it is true, because rerunning one configuration changes the loss by more than most configuration changes do; on Chip-Bench, an experiment called an improvement beats the run's best no more often than one the agent said nothing about.
The agents act on these verdicts (\autoref{tab:verdicts}, right): on the speedrun, \btSpeedFalseWinKidsPct\% of the follow-up experiments (\btSpeedFalseWinKids{} of \btSpeedKids{}) build on a win that was not real. An agent that chooses its next experiments from its own verdicts on noisy results is responding partly to noise.

\subsection{Late-stage exploration distinguishes winning runs}
\label{sec:exp-winners}

\begin{figure}[!tbp]
\centering
\includegraphics[width=\linewidth]{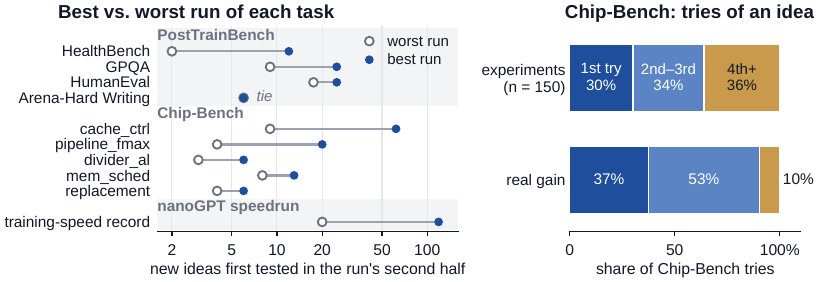}
\caption{\textbf{Winning runs keep testing new ideas late, and an idea's later tries add little.} Left: new ideas first tested in the second half of the best and the worst run of each task. Right: on Chip-Bench, an idea's fourth and later tries are \lateTryExpPct\% of experiments but bring \lateTryGainPct\% of the real gain.}
\label{fig:winners}
\end{figure}

We ranked the runs of each task by outcome and asked which statistics of the agent's actions alone go with a better rank.
Run length does not matter: steps and outcome are unrelated (rank correlation \rhoStepsAll{} over \nFindRuns{} runs).
What the steps are spent on does.
The statistic that most strongly separates winners from losers is the number of ideas first tested in the second half of the run (\rhoLate{}, positive on each benchmark). The best run of a task tests more such ideas than the worst on \lateUp{} of \lateTasks{} tasks (\autoref{fig:winners}, left).
Dividing the count by the number of steps leaves the correlation at \rhoLatePerStep{}, and when one model ran the same task twice, the higher-scoring run tested more such ideas in \latePairUp{} of \latePairN{} pairs.

The logged results explain why breadth pays.
On Chip-Bench, an idea's fourth and later tries are \lateTryExpPct\% of experiments and win about as often as its first three (\lateTryWinPct\% against \earlyTryWinPct\%), but they contribute only \lateTryGainPct\% of the real gain, and the gap holds within each third of a run, so it does not come from when the tries happen (\autoref{fig:winners}, right).
On the speedrun ideas rarely reach a fourth try, and first tries bring \speedFirstGainPct\% of the gain (\nTruthChip{} Chip-Bench and \nTruthSpeed{} speedrun runs log these results).

Testing more ideas takes time, yet no Chip-Bench run used its full time budget: the median stopped at \chipTimeUsedMedian\% of it, and on every task the best run stopped later than the worst.

\subsection{The task decides whether agents hack; the model decides how}
\label{sec:exp-hack}

\begin{table}[t]
\centering\small
\begin{minipage}[t]{0.47\linewidth}\vspace{0pt}\centering\begin{tabular}{@{}l@{\hspace{6pt}}c@{\hspace{6pt}}c@{\hspace{6pt}}c@{}}
\toprule
 & \multicolumn{2}{c}{\textbf{E4}} & \textbf{Auditors:} \\
\cmidrule(lr){2-3}
\textbf{Task} & \textbf{flagged} & \textbf{per run} & \textbf{cheated} \\
\midrule
BFCL & 10/14 & 2.29 & 8/14 \\
HumanEval & 3/12 & 0.25 & 0/11 \\
HealthBench & 5/20 & 0.25 & 1/20 \\
GPQA & 2/10 & 0.20 & 0/10 \\
Arena-Hard & 3/20 & 0.15 & 2/20 \\
AIME & 0/15 & 0.00 & 0/15 \\
\bottomrule
\end{tabular}
\end{minipage}\hfill
\begin{minipage}[t]{0.47\linewidth}\vspace{0pt}\centering\begin{tabular}{@{}l@{\hspace{6pt}}c@{\hspace{6pt}}l@{}}
\toprule
 & \multicolumn{2}{c}{\textbf{E4}} \\
\cmidrule(l){2-3}
\textbf{Model} & \textbf{per run} & \textbf{main method} \\
\midrule
GLM-5.2 & 1.6 & data from test items \\
Kimi K3 & 0.8 & data from test items \\
Claude Fable 5 & 0.5 & fits test format \\
Claude Opus 5 & 0.4 & data from test items \\
GPT-5.6 & 0.3 & luckiest rerun \\
GPT-5.5 & 0.1 & luckiest rerun \\
\bottomrule
\end{tabular}
\end{minipage}
\caption{\textbf{The task decides whether agents hack; the model decides how.} Left: E4's reward-hacking instances per PostTrainBench run, by task, beside the auditors' rulings. Right: each model's rate and most frequent method on the tasks where some run hacked.}
\label{tab:hack}
\end{table}

\begin{figure}[!tbp]
\centering
\includegraphics[width=\linewidth]{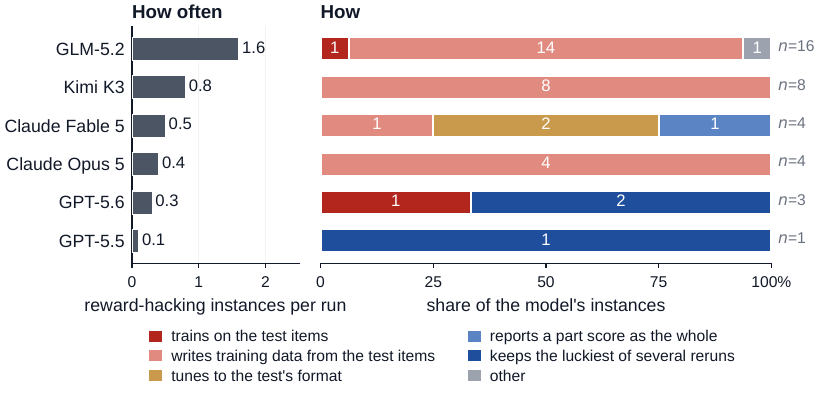}
\caption{\textbf{Agent models deploy distinct reward-hacking strategies when evaluation shortcuts exist.} Rather than cheating identically, different model families exploit specific loopholes (such as synthesizing training data from test failures or fishing for favorable seeds).}
\label{fig:hack-methods}
\end{figure}

Reward hacking needs an opening, and the tasks differ in how open they are (\autoref{tab:hack}, left): on BFCL, E4 flags most runs and the auditors agree on most of them, while on AIME 2025 no run of any agent hacked.
Where an opening exists, models differ in how they hack as well as how often (\autoref{tab:hack}, right).
On the \hmTasks{} tasks where at least one agent hacked, the judge identifies the broken rule and quotes supporting record text in its response to E4's query for each instance; we group the \hmIdeas{} instances of six models by that rule in \autoref{fig:hack-methods}.
GLM-5.2 and Kimi K3 hack most (\hmRateGLM{} and \hmRateKimi{} instances per run) and almost always the same way: \hmDeriveGLMKimi{} of their \hmIdeasGLMKimi{} instances involve training data written from test items, where the agent evaluates, reads which test items failed, writes examples that cover them, and evaluates again.
The GPT models hack rarely (\hmRateGPTsix{} and \hmRateGPTfive{}) and differently: they seldom touch test data, and in \hmLuckyGPT{} of their \hmIdeasGPT{} instances, the agents rerun one recipe under other seeds and keep the best-scoring draw.
Claude Opus 5 hacks like GLM-5.2 but far less often (\hmDeriveOpus{} of \hmIdeasOpus{}), and Claude Fable 5 has the widest repertoire, among it tuning to the test's answer format and reporting a score on part of the test as the score on the whole.
A benchmark's audit answers whether a run cheated; the record identifies the actions involved.

\subsection{The persona comes with the model}
\label{sec:exp-model}

\begin{table}[t]
\centering\small
\resizebox{\linewidth}{!}{\begin{tabular}{@{}lccc@{\hspace{14pt}}cccccc@{}}
\toprule
 & \multicolumn{3}{c}{\textbf{Variance explained by}} & \multicolumn{6}{c}{\textbf{Mean score of each model}} \\
\cmidrule(lr){2-4}\cmidrule(l){5-10}
\textbf{Trait} & \textbf{model} & \textbf{task} & \textbf{chance} & \textbf{Opus 5} & \textbf{Fable 5} & \textbf{GPT-5.6} & \textbf{GPT-5.5} & \textbf{Kimi K3} & \textbf{GLM-5.2} \\
\midrule
T1 budget & \textbf{75\%} & 3\% & 13\% & .25 & .22 & .10 & .10 & .33 & .23 \\
T2 concurrency & \textbf{32\%} & 8\% & 13\% & .44 & .49 & .27 & .24 & .28 & .27 \\
T3 search & \textbf{36\%} & 6\% & 15\% & .40 & .39 & .16 & .28 & .26 & .27 \\
T4 failure style & \textbf{31\%} & 16\% & 21\% & -- & -- & .52 & .40 & .44 & .52 \\
T5 provenance & \textbf{51\%} & 18\% & 35\% & -- & -- & .41 & -- & -- & -- \\
T6 tempo & \textbf{59\%} & 2\% & 17\% & .02 & .03 & .33 & .43 & .12 & .41 \\
\bottomrule
\end{tabular}
}
\caption{\textbf{The model sets the persona.} For each trait over up to \nPersonaRuns{} PostTrainBench runs: the variance the model and the task explain, the model's share by chance, and each model's mean score.}
\label{tab:persona}
\end{table}

\begin{figure}[!tbp]
\centering
\includegraphics[width=\linewidth]{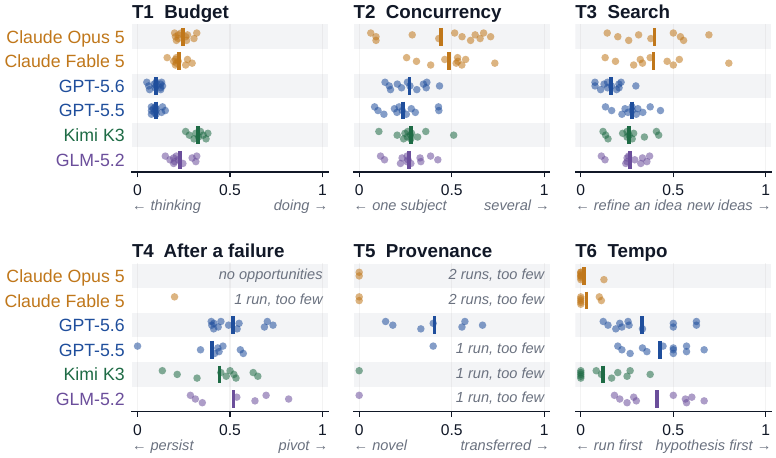}
\caption{\textbf{Persona reflects model identity rather than the task.} Models exhibit consistent, distinct research habits across tasks, maintaining characteristic preferences for budgeting action versus reasoning and for stating hypotheses before execution.}
\label{fig:persona}
\end{figure}

For every persona trait, the model explains more of the variance than the task and more than chance (\autoref{tab:persona}, left).\footnote{We use two-factor least squares, crediting each factor only with the variance it explains after accounting for the other. As a chance baseline, we report the share explained after shuffling that factor's labels.}
The profiles differ by model (\autoref{tab:persona}, right): the two Claude models almost never state a hypothesis before the experiment that tests it (T6), while GPT-5.5, GPT-5.6 and GLM-5.2 state a third or more of their propositions first; the GPT models spend the smallest share of their output on actions (T1). \autoref{fig:persona} shows every run.
A harness should therefore work with each model's research habits and target safeguards to each model.

\section{Related work}
\label{sec:related}

\paragraph{Outcome benchmarks.}
Agent benchmarks score ML engineering \citep{chan2024mlebench,chen2025mlrbench,starace2025paperbench}, scientific discovery \citep{chen2024scienceagentbench,lu2024aiscientist}, and software engineering \citep{jimenez2024swebench,yao2025taubench} by unit tests, predictive performance, or reproduced results; ALE-Bench scores open-ended optimization without a known optimum \citep{imajuku2025alebench}.
Such scores do not establish whether an agent's claims are supported by its experiments, and up to half of their variance can stem from the harness \citep{taubenchaudit2026}. \oeb{} complements them and scores their logs once converted into its record format.

\paragraph{Process evaluation.}
Step-level labels train better reward models for mathematical solutions \citep{lightman2023verify}. AgentBoard tracks progress on annotated subgoals \citep{ma2024agentboard}, PaperBench grades replication against expert rubrics that took weeks per paper \citep{starace2025paperbench}, DiscoveryWorld scores procedures with task-specific scorecards \citep{jansen2024discoveryworld}, and AgentRx locates execution errors \citep{agentrx2026}.
These methods need task-specific targets; \oeb{} derives its opportunities from structural rules of the scientific method and needs no task rubric.

\paragraph{LLM judges.}
LLM judges agree with human preferences on text \citep{zheng2023judging}, but in a synthetic case study they rated an incorrect claim highly when shown only the agent's prose answer \citep{groundeval2026}, and they struggle to locate decisive errors in multi-agent logs \citep{zhang2025whowhen}.
\oeb{} therefore asks judges only closed questions about verified excerpts and leaves assembly and counting to code.

\paragraph{Open-ended discovery.}
Open-endedness measures novelty \citep{wang2020enhancedpoet}, interestingness \citep{zhang2023omni}, state coverage \citep{crafter2025}, or world-model accuracy \citep{warrier2026benchmarking,vafa2024worldmodel}. \oeb{} checks that commitments match evidence from its actions.

\FloatBarrier  %
\section{Limitations and conclusion}
\label{sec:limits}

\paragraph{Limitations.}
\begin{itemize}[leftmargin=1.2em,itemsep=1pt,topsep=2pt]
  \item \emph{Inference expressed through numbers.} In pure hyperparameter search, models often reason only through the values they change, so the instrument extracts fewer propositions.
  \item \emph{Reasoning off the record.} E1--E3 audit the account the agent writes. When that account is missing, the instrument cannot tell unwritten reasoning from reasoning that did not occur, so the outcome analyses of \autoref{sec:exp-winners} read only executed acts.
  \item \emph{Optimizing against the instrument.} The runs studied here were produced without knowledge of \oeb{}. How far an agent trained or prompted against its scores can move them is the natural next stress test of the defenses in \autoref{sec:problem}.
  \item \emph{Computational cost.} A run took a median of \callsMedian{} judge calls, which suits offline evaluation rather than a training reward.
\end{itemize}

\paragraph{Conclusion.}
We have presented \oeb{}, a benchmark-agnostic methodology that measures epistemic process from agent execution records alone.
\oeb{} compiles records into a unified epistemic event graph under the principle that \emph{semantics proposes, actions dispose}: stated propositions are linked to the executed actions that test them. It scores epistemic competence without outcome grading or human-written rubrics, and its scores hold under rescoring (median change \conSameMedian{}).
The records also show that only \vtSpeedGoodPct--\vtChipGoodPct\% of claimed improvements are real, that winning runs keep testing new ideas late (\lateUp{} of \lateTasks{} tasks), and that the model explains more of each trait's variance than the task (median \varTraitModelMedian\% vs.\ \varTraitTaskMedian\%). Building harnesses on these readings, and testing them against equal-cost baselines, is the next step.

\subsubsection*{Reproducibility Statement}
The evaluation code (the record format, the rules for converting a benchmark's logs into it (\autoref{sec:app-conversion}), every judge prompt, the graph compiler, and the measurement layer) is released under an MIT license at \url{https://github.com/ARA-Labs/oeb}, together with the converters for the three benchmarks.
For every scored run reported here, a separate dataset (\url{https://huggingface.co/datasets/AgentNativeResearchLab/oeb-scored-runs}) holds the converted record, the graph, each jury's output, the complete log of judge requests and responses (which can be replayed), and the scores, including the persona trait values from which the per-model distributions follow. We generate every table, figure, and number in \autoref{sec:validity} and \autoref{sec:exp} from those files with the scripts in the code repository.

\subsubsection*{Ethics Statement}
This research evaluates autonomous AI agents using existing computational benchmarks.
No human subjects or private personal data were involved.
We discuss the remaining risk that agents could raise their scores without improving their research behavior in \autoref{sec:problem} and \autoref{sec:app-attacks}.

\subsubsection*{AI Use Statement}
In this work, we used generative AI tools to implement methods (the conversion, graph compiler, and measurement code, and the analysis scripts), to give feedback on the research methodology, and to support the analysis and interpretation of results. A language model is also a component of the evaluation method itself: it extracts cards and answers closed jury questions (\autoref{sec:method}), and code verifies every excerpt it quotes against the source record, which establishes only that the text was present.
We have not used generative AI tools to generate synthetic data or to propose the research hypotheses; formulating and proving mathematical claims is not applicable to this work.
Additionally, we used generative AI tools to draft and edit the paper and to create figures. We have reviewed all AI-assisted work: the authors audited scored records by hand against the raw logs and checked every reported number against the generating scripts. We take responsibility for the final content of this work, including text, claims, and artifacts produced with the aid of generative AI.

\bibliographystyle{plainnat}
\bibliography{refs}

\begin{thebibliography}{28}
\providecommand{\natexlab}[1]{#1}
\providecommand{\url}[1]{\texttt{#1}}
\expandafter\ifx\csname urlstyle\endcsname\relax
  \providecommand{\doi}[1]{doi: #1}\else
  \providecommand{\doi}{doi: \begingroup \urlstyle{rm}\Url}\fi

\bibitem[{ARA Labs}(2026)]{chipbench2026}
{ARA Labs}.
\newblock {Chip-Bench: AI Chip Design Benchmark}, 2026.
\newblock URL \url{https://www.chipbench.ai/}.
\newblock Accessed 2026-09-25.

\bibitem[Barke et~al.(2026)Barke, Goyal, Khare, Singh, Nath, and
  Bansal]{agentrx2026}
Shraddha Barke, Arnav Goyal, Alind Khare, Avaljot Singh, Suman Nath, and Chetan
  Bansal.
\newblock {AgentRx: Diagnosing AI Agent Failures from Execution Trajectories},
  2026.
\newblock URL \url{https://arxiv.org/abs/2602.02475}.

\bibitem[Chan et~al.(2025)Chan, Chowdhury, Jaffe, Aung, Sherburn, Mays,
  Starace, Liu, Maksin, Patwardhan, Weng, and M{\k{a}}dry]{chan2024mlebench}
Jun~Shern Chan, Neil Chowdhury, Oliver Jaffe, James Aung, Dane Sherburn, Evan
  Mays, Giulio Starace, Kevin Liu, Leon Maksin, Tejal Patwardhan, Lilian Weng,
  and Aleksander M{\k{a}}dry.
\newblock {MLE-bench: Evaluating Machine Learning Agents on Machine Learning
  Engineering}.
\newblock In \emph{The Thirteenth International Conference on Learning
  Representations (ICLR)}, 2025.
\newblock URL \url{https://openreview.net/forum?id=6s5uXNWGIh}.

\bibitem[Chen et~al.(2025{\natexlab{a}})Chen, Xiong, Lu, Han, Deng, He, Wu, Li,
  Liu, and Hooi]{chen2025mlrbench}
Hui Chen, Miao Xiong, Yujie Lu, Wei Han, Ailin Deng, Yufei He, Jiaying Wu, Yibo
  Li, Yue Liu, and Bryan Hooi.
\newblock {MLR-Bench: Evaluating AI Agents on Open-Ended Machine Learning
  Research}.
\newblock In \emph{Advances in Neural Information Processing Systems 38
  (NeurIPS 2025)}, pages 131093--131141. Neural Information Processing Systems
  Foundation, Inc. (NeurIPS), 2025{\natexlab{a}}.
\newblock \doi{10.52202/085713-3945}.

\bibitem[Chen et~al.(2025{\natexlab{b}})Chen, Chen, Ning, Zhang, Wang, Yu, Li,
  Liao, Wei, Lu, Dey, Xue, Baker, Burns, Adu-Ampratwum, Huang, Ning, Gao, Su,
  and Sun]{chen2024scienceagentbench}
Ziru Chen, Shijie Chen, Yuting Ning, Qianheng Zhang, Boshi Wang, Botao Yu,
  Yifei Li, Zeyi Liao, Chen Wei, Zitong Lu, Vishal Dey, Mingyi Xue, Frazier~N.
  Baker, Benjamin Burns, Daniel Adu-Ampratwum, Xuhui Huang, Xia Ning, Song Gao,
  Yu~Su, and Huan Sun.
\newblock {ScienceAgentBench: Toward Rigorous Assessment of Language Agents for
  Data-Driven Scientific Discovery}.
\newblock In \emph{The Thirteenth International Conference on Learning
  Representations (ICLR)}, 2025{\natexlab{b}}.
\newblock URL \url{https://openreview.net/forum?id=6z4YKr0GK6}.

\bibitem[Flynt(2026)]{groundeval2026}
Jeffrey Flynt.
\newblock {GroundEval: A Deterministic Replacement for LLM-as-Judge in Stateful
  Agent Evaluation}, 2026.
\newblock URL \url{https://arxiv.org/abs/2606.22737}.

\bibitem[Imajuku et~al.(2025)Imajuku, Horie, Iwata, Aoki, Takahashi, and
  Akiba]{imajuku2025alebench}
Yuki Imajuku, Kohki Horie, Yoichi Iwata, Kensho Aoki, Naohiro Takahashi, and
  Takuya Akiba.
\newblock {ALE-Bench: A Benchmark for Long-Horizon Objective-Driven Algorithm
  Engineering}.
\newblock In \emph{Advances in Neural Information Processing Systems 38
  (NeurIPS 2025)}, pages 173630--173675. Neural Information Processing Systems
  Foundation, Inc. (NeurIPS), 2025.
\newblock \doi{10.52202/085713-5226}.

\bibitem[Jansen et~al.(2024)Jansen, Côté, Khot, Bransom, Mishra, Majumder,
  Tafjord, and Clark]{jansen2024discoveryworld}
Peter Jansen, Marc-Alexandre Côté, Tushar Khot, Erin Bransom, Bhavana~Dalvi
  Mishra, Bodhisattwa~Prasad Majumder, Oyvind Tafjord, and Peter Clark.
\newblock {DiscoveryWorld: A Virtual Environment for Developing and Evaluating
  Automated Scientific Discovery Agents}.
\newblock In \emph{Advances in Neural Information Processing Systems 37
  (NeurIPS 2024)}, pages 10088--10116. Neural Information Processing Systems
  Foundation, Inc. (NeurIPS), 2024.
\newblock \doi{10.52202/079017-0324}.

\bibitem[Jimenez et~al.(2024)Jimenez, Yang, Wettig, Yao, Pei, Press, and
  Narasimhan]{jimenez2024swebench}
Carlos~E. Jimenez, John Yang, Alexander Wettig, Shunyu Yao, Kexin Pei, Ofir
  Press, and Karthik Narasimhan.
\newblock {SWE-bench: Can Language Models Resolve Real-World GitHub Issues?}
\newblock In \emph{The Twelfth International Conference on Learning
  Representations (ICLR)}, 2024.
\newblock URL \url{https://openreview.net/forum?id=VTF8yNQM66}.

\bibitem[Kirgis et~al.(2026)Kirgis, Kapoor, Rabanser, Nadgir, Ududec, Dubois,
  Allaire, Stosz, Hobbhahn, Steinhardt, and Narayanan]{taubenchaudit2026}
Peter Kirgis, Sayash Kapoor, Stephan Rabanser, Nitya Nadgir, Cozmin Ududec,
  Magda Dubois, JJ~Allaire, Conrad Stosz, Marius Hobbhahn, Jacob Steinhardt,
  and Arvind Narayanan.
\newblock {Log analysis is necessary for credible evaluation of AI agents},
  2026.
\newblock URL \url{https://arxiv.org/abs/2605.08545}.

\bibitem[Klahr and Dunbar(1988)]{klahr1988dual}
David Klahr and Kevin Dunbar.
\newblock Dual space search during scientific reasoning.
\newblock \emph{Cognitive Science}, 12\penalty0 (1):\penalty0 1--48, 1988.
\newblock \doi{10.1207/s15516709cog1201_1}.
\newblock URL \url{https://doi.org/10.1207/s15516709cog1201_1}.

\bibitem[Lidayan et~al.(2025)Lidayan, Du, Kosoy, Rufova, Abbeel, and
  Gopnik]{crafter2025}
Aly Lidayan, Yuqing Du, Eliza Kosoy, Maria Rufova, Pieter Abbeel, and Alison
  Gopnik.
\newblock {Intrinsically-Motivated Humans and Agents in Open-World
  Exploration}, 2025.
\newblock URL \url{https://arxiv.org/abs/2503.23631}.

\bibitem[Lightman et~al.(2024)Lightman, Kosaraju, Burda, Edwards, Baker, Lee,
  Leike, Schulman, Sutskever, and Cobbe]{lightman2023verify}
Hunter Lightman, Vineet Kosaraju, Yura Burda, Harri Edwards, Bowen Baker, Teddy
  Lee, Jan Leike, John Schulman, Ilya Sutskever, and Karl Cobbe.
\newblock {Let's Verify Step by Step}.
\newblock In \emph{The Twelfth International Conference on Learning
  Representations (ICLR)}, 2024.
\newblock URL \url{https://openreview.net/forum?id=v8L0pN6EOi}.

\bibitem[Lu et~al.(2024)Lu, Lu, Lange, Foerster, Clune, and
  Ha]{lu2024aiscientist}
Chris Lu, Cong Lu, Robert~Tjarko Lange, Jakob Foerster, Jeff Clune, and David
  Ha.
\newblock {The AI Scientist: Towards Fully Automated Open-Ended Scientific
  Discovery}, 2024.
\newblock URL \url{https://arxiv.org/abs/2408.06292}.

\bibitem[Luo et~al.(2025)Luo, Kasirzadeh, and Shah]{pitfalls2025}
Ziming Luo, Atoosa Kasirzadeh, and Nihar~B. Shah.
\newblock {The More You Automate, the Less You See: Hidden Pitfalls of AI
  Scientist Systems}, 2025.
\newblock URL \url{https://arxiv.org/abs/2509.08713}.

\bibitem[Ma et~al.(2024)Ma, Zhang, Zhu, Yang, Yang, Jin, Lan, Kong, and
  He]{ma2024agentboard}
Chang Ma, Junlei Zhang, Zhihao Zhu, Cheng Yang, Yujiu Yang, Yaohui Jin,
  Zhenzhong Lan, Lingpeng Kong, and Junxian He.
\newblock {AgentBoard: An Analytical Evaluation Board of Multi-turn LLM
  Agents}.
\newblock In \emph{Advances in Neural Information Processing Systems 37
  (NeurIPS 2024)}, pages 74325--74362. Neural Information Processing Systems
  Foundation, Inc. (NeurIPS), 2024.
\newblock \doi{10.52202/079017-2365}.

\bibitem[{Prime Intellect}(2026)]{primeintellect2026speedrun}
{Prime Intellect}.
\newblock {NanoGPT Speedrun}, 2026.
\newblock URL \url{https://app.primeintellect.ai/speedrun/nanogpt}.
\newblock Accessed 2026-09-25.

\bibitem[Rank et~al.(2026)Rank, Bhatnagar, Prabhu, Eisenberg, Nguyen, Bethge,
  and Andriushchenko]{rank2026posttrainbench}
Ben Rank, Hardik Bhatnagar, Ameya Prabhu, Shira Eisenberg, Karina Nguyen,
  Matthias Bethge, and Maksym Andriushchenko.
\newblock {PostTrainBench: Can LLM Agents Automate LLM Post-Training?}, 2026.
\newblock URL \url{https://arxiv.org/abs/2603.08640}.

\bibitem[Starace et~al.(2025)Starace, Jaffe, Sherburn, Aung, Chan, Maksin,
  Dias, Mays, Kinsella, Thompson, Heidecke, Glaese, and
  Patwardhan]{starace2025paperbench}
Giulio Starace, Oliver Jaffe, Dane Sherburn, James Aung, Jun~Shern Chan, Leon
  Maksin, Rachel Dias, Evan Mays, Benjamin Kinsella, Wyatt Thompson, Johannes
  Heidecke, Amelia Glaese, and Tejal Patwardhan.
\newblock {PaperBench: Evaluating AI's Ability to Replicate AI Research}.
\newblock In \emph{Proceedings of the 42nd International Conference on Machine
  Learning}, volume 267 of \emph{Proceedings of Machine Learning Research},
  pages 56843--56873. PMLR, 2025.
\newblock URL \url{https://proceedings.mlr.press/v267/starace25a.html}.

\bibitem[Vafa et~al.(2024)Vafa, Chen, Rambachan, Kleinberg, and
  Mullainathan]{vafa2024worldmodel}
Keyon Vafa, Justin Chen, Ashesh Rambachan, Jon Kleinberg, and Sendhil
  Mullainathan.
\newblock {Evaluating the World Model Implicit in a Generative Model}.
\newblock In \emph{Advances in Neural Information Processing Systems 37
  (NeurIPS 2024)}, pages 26941--26975. Neural Information Processing Systems
  Foundation, Inc. (NeurIPS), 2024.
\newblock \doi{10.52202/079017-0846}.

\bibitem[Verga et~al.(2024)Verga, Hofstatter, Althammer, Su, Piktus,
  Arkhangorodsky, Xu, White, and Lewis]{verga2024juries}
Pat Verga, Sebastian Hofstatter, Sophia Althammer, Yixuan Su, Aleksandra
  Piktus, Arkady Arkhangorodsky, Minjie Xu, Naomi White, and Patrick Lewis.
\newblock {Replacing Judges with Juries: Evaluating LLM Generations with a
  Panel of Diverse Models}, 2024.
\newblock URL \url{https://arxiv.org/abs/2404.18796}.

\bibitem[Wang et~al.(2020)Wang, Lehman, Rawal, Zhi, Li, Clune, and
  Stanley]{wang2020enhancedpoet}
Rui Wang, Joel Lehman, Aditya Rawal, Jiale Zhi, Yulun Li, Jeffrey Clune, and
  Kenneth Stanley.
\newblock {Enhanced POET: Open-ended Reinforcement Learning through Unbounded
  Invention of Learning Challenges and their Solutions}.
\newblock In \emph{Proceedings of the 37th International Conference on Machine
  Learning}, volume 119 of \emph{Proceedings of Machine Learning Research},
  pages 9940--9951. PMLR, 2020.
\newblock URL \url{https://proceedings.mlr.press/v119/wang20l.html}.

\bibitem[Warrier et~al.(2026)Warrier, Nguyen, Naim, Jain, Liang, Schroeder,
  Yang, Tenenbaum, Vollmer, Ellis, and Tavares]{warrier2026benchmarking}
Archana Warrier, Dat Nguyen, Michelangelo Naim, Moksh Jain, Yichao Liang, Karen
  Schroeder, Cambridge Yang, Joshua~B. Tenenbaum, Sebastian~Josef Vollmer,
  Kevin Ellis, and Zenna Tavares.
\newblock Benchmarking world-model learning with environment-level queries.
\newblock In \emph{Forty-third International Conference on Machine Learning},
  2026.
\newblock URL \url{https://openreview.net/forum?id=Ny6UYZ8ysN}.

\bibitem[Wijk et~al.(2025)Wijk, Lin, Becker, Jawhar, Parikh, Broadley, Chan,
  Chen, Clymer, Dhyani, Ericheva, Garcia, Goodrich, Jurkovic, Kinniment, Lajko,
  Nix, Koba~Sato, Saunders, Taran, West, and Barnes]{wijk2024rebench}
Hjalmar Wijk, Tao~Roa Lin, Joel Becker, Sami Jawhar, Neev Parikh, Thomas
  Broadley, Lawrence Chan, Michael Chen, Joshua~M Clymer, Jai Dhyani, Elena
  Ericheva, Katharyn Garcia, Brian Goodrich, Nikola Jurkovic, Megan Kinniment,
  Aron Lajko, Seraphina Nix, Lucas~Jun Koba~Sato, William Saunders, Maksym
  Taran, Ben West, and Elizabeth Barnes.
\newblock {RE}-bench: Evaluating frontier {AI} r\&d capabilities of language
  model agents against human experts.
\newblock In \emph{Proceedings of the 42nd International Conference on Machine
  Learning}, volume 267 of \emph{Proceedings of Machine Learning Research},
  pages 66772--66832. PMLR, 13--19 Jul 2025.
\newblock URL \url{https://proceedings.mlr.press/v267/wijk25a.html}.

\bibitem[Yao et~al.(2025)Yao, Shinn, Razavi, and Narasimhan]{yao2025taubench}
Shunyu Yao, Noah Shinn, Pedram Razavi, and Karthik~R Narasimhan.
\newblock {$\tau$-bench: A Benchmark for Tool-Agent-User Interaction in
  Real-World Domains}.
\newblock In \emph{The Thirteenth International Conference on Learning
  Representations}, 2025.
\newblock URL \url{https://openreview.net/forum?id=roNSXZpUDN}.

\bibitem[Zhang et~al.(2024)Zhang, Lehman, Stanley, and Clune]{zhang2023omni}
Jenny Zhang, Joel Lehman, Kenneth Stanley, and Jeff Clune.
\newblock {OMNI: Open-endedness via Models of human Notions of
  Interestingness}.
\newblock In \emph{The Twelfth International Conference on Learning
  Representations (ICLR)}, 2024.
\newblock URL \url{https://openreview.net/forum?id=AgM3MzT99c}.

\bibitem[Zhang et~al.(2025)Zhang, Yin, Zhang, Liu, Han, Zhang, Li, Wang, Wang,
  Chen, and Wu]{zhang2025whowhen}
Shaokun Zhang, Ming Yin, Jieyu Zhang, Jiale Liu, Zhiguang Han, Jingyang Zhang,
  Beibin Li, Chi Wang, Huazheng Wang, Yiran Chen, and Qingyun Wu.
\newblock {Which Agent Causes Task Failures and When? On Automated Failure
  Attribution of LLM Multi-Agent Systems}.
\newblock In \emph{Proceedings of the 42nd International Conference on Machine
  Learning}, volume 267 of \emph{Proceedings of Machine Learning Research},
  pages 76583--76599. PMLR, 2025.
\newblock URL \url{https://proceedings.mlr.press/v267/zhang25cq.html}.

\bibitem[Zheng et~al.(2023)Zheng, Chiang, Sheng, Zhuang, Wu, Zhuang, Lin, Li,
  Li, Xing, Zhang, Gonzalez, and Stoica]{zheng2023judging}
Lianmin Zheng, Wei-Lin Chiang, Ying Sheng, Siyuan Zhuang, Zhanghao Wu, Yonghao
  Zhuang, Zi~Lin, Zhuohan Li, Dacheng Li, Eric Xing, Hao Zhang, Joseph
  Gonzalez, and Ion Stoica.
\newblock {Judging LLM-as-a-Judge with MT-Bench and Chatbot Arena}.
\newblock In \emph{Advances in Neural Information Processing Systems 36
  (NeurIPS 2023)}, pages 46595--46623. Neural Information Processing Systems
  Foundation, Inc. (NeurIPS), 2023.
\newblock \doi{10.52202/075280-2020}.

\end{thebibliography}

\clearpage
\appendix
\section*{Appendix}
\renewcommand{\thesubsection}{\Alph{subsection}}
\renewcommand{\theHsubsection}{appendix.\Alph{subsection}}

\subsection{Five ways to farm the rubric and the defense against each}
\label{sec:app-attacks}
An agent that knows the rubric could write its record in five ways to game the evaluation; each entry names the defense the instrument is designed with:
\emph{A1 (cheap declaration)}: stating high volumes of trivial hypotheses; designed against by crediting no proposition on its own: every count that credits the agent requires an executed act linked to the proposition, and a belief asserted before any observation bears on it lowers \emph{belief grounded} (E1).
\emph{A2 (observation parroting)}: presenting information from tool observations as newly generated insights; designed against by verifying when the information first appeared in the record.
\emph{A3 (self-citation farming)}: citing prior thoughts as evidence; designed against by requiring graph edges that connect stated propositions to executed actions.
\emph{A4 (ritual acknowledgment)}: conceding errors verbally without altering code; designed against by requiring subsequent revisions to the executed work.
\emph{A5 (collision avoidance)}: refusing to state falsifiable predictions; penalized because every launched experiment enters the denominator of experiment yield (E2), whether or not the agent stated what it expected.

\subsection{L0 conversion rules}
\label{sec:app-conversion}
Converting a benchmark's logs (L0) must:
\begin{enumerate*}[label=(\roman*)]
  \item divide the raw log into steps deterministically;
  \item tag each step as an environment, workspace, or submission step;
  \item extract action and observation strings without changing their text; and
  \item record in the world manifest (\texttt{world.json}) the evaluation rules and setup the log does not state, such as the task's rules and the size of its evaluation set.
\end{enumerate*}
The conversion needs no knowledge of the downstream scoring rules.

\subsection{Prompt specifications}
\label{sec:app-prompts}
Extraction and jury prompts specify the types of card to identify and require quoted excerpts to occur exactly in the source text.
Each extraction prompt reads a five-step window of the record; each jury prompt asks one closed question about one card or a pair of cards. Three independent judge calls answer the question; whenever a vote quotes the record, code verifies that the excerpt occurs verbatim in the source.
Prompts list neither examples of qualifying content nor their own audit history.

\subsection{Reliability details}
\label{app:reliability}

For each dimension, we report the mean absolute score difference when the same record is scored twice, either by the same judge or by two different judges (\autoref{tab:reliability}). We also compare scores from two independent runs of the same agent on the same task and scores from different agent models.
We report the per-model means of the competence axes over all runs of the six PostTrainBench tasks, each of which post-trains gemma-3-4b, in \autoref{tab:main}.
Across the constructs of E1--E3 and T2--T6, rescoring by the same judge changed a numerator count by a median of \conSameCount{} (\conOtherCount{} under the other judge). The construct with the most opportunities had a median of \conSameDenomMax{} per run.

\begin{table}[!htbp]
\centering\small
\resizebox{\linewidth}{!}{\begin{tabular}{@{}lcccccccccc@{}}
\toprule
 & \textbf{E1} & \textbf{E2} & \textbf{E3} & \textbf{E4} & \textbf{T1} & \textbf{T2} & \textbf{T3} & \textbf{T4} & \textbf{T5} & \textbf{T6} \\
\midrule
Same record, judged twice & 0.04\,\textsubscript{37} & 0.05\,\textsubscript{40} & 0.16\,\textsubscript{12} & 0.09\,\textsubscript{40} & 0.00\,\textsubscript{35} & 0.09\,\textsubscript{39} & 0.13\,\textsubscript{35} & 0.11\,\textsubscript{21} & 0.13\,\textsubscript{4} & 0.10\,\textsubscript{28} \\
Same agent, second run & 0.12\,\textsubscript{33} & 0.14\,\textsubscript{35} & 0.23\,\textsubscript{10} & 0.14\,\textsubscript{35} & 0.02\,\textsubscript{31} & 0.16\,\textsubscript{35} & 0.11\,\textsubscript{31} & 0.13\,\textsubscript{15} & 0.10\,\textsubscript{2} & 0.16\,\textsubscript{27} \\
Different agent models & 0.15\,\textsubscript{80} & 0.23\,\textsubscript{85} & 0.25\,\textsubscript{13} & 0.17\,\textsubscript{85} & 0.11\,\textsubscript{71} & 0.18\,\textsubscript{85} & 0.15\,\textsubscript{71} & 0.14\,\textsubscript{17} & 0.33\,\textsubscript{1} & 0.28\,\textsubscript{59} \\
\midrule
Same record, other judge$^\dagger$ & 0.16\,\textsubscript{10} & 0.08\,\textsubscript{12} & 0.17\,\textsubscript{2} & 0.00\,\textsubscript{12} & 0.07\,\textsubscript{11} & 0.08\,\textsubscript{12} & 0.14\,\textsubscript{12} & 0.09\,\textsubscript{5} & 0.36\,\textsubscript{2} & 0.09\,\textsubscript{11} \\
\bottomrule
\end{tabular}
}
\caption{\textbf{Scoring variation across repeated evaluations, independent runs, and different agents.} Each cell is the mean absolute score difference; subscripts count the score pairs compared, where a pair counts only if both scores exist. Scores remain stable when the same judge rescores a record and consistent when a different judge does. $^\dagger$12 Chip-Bench records scored by GLM-5.3, the judge for PostTrainBench and Chip-Bench, and again by GPT-5.6; the other rows use PostTrainBench records.}
\label{tab:reliability}
\end{table}

\end{document}